\pdfoutput=1
\documentclass[11pt]{article}
\usepackage[table,xcdraw,dvipsnames]{xcolor}
\usepackage{enumerate}
\usepackage{times}
\usepackage{expex}
\usepackage{latexsym}
\usepackage[T1]{fontenc}
\usepackage[utf8]{inputenc}
\definecolor{darkorange}{rgb}{1, 0.549, 0}

\usepackage{microtype}
\usepackage{enumitem} 
\usepackage{bbding}
\usepackage{CJKutf8}
\usepackage{graphicx}
\usepackage{array}
\usepackage{arydshln}
\usepackage{multirow}

\usepackage{bm}
\usepackage{makecell}
\usepackage{wrapfig}
\usepackage{lipsum}  
\usepackage{etoolbox}
\usepackage[final]{acl2023}
\usepackage{booktabs}
  
\usepackage{csquotes}
\usepackage{bm}
\usepackage{amsmath,amssymb, amsfonts,mathtools}
\usepackage{soul}
\usepackage{adjustbox}
\usepackage[]{todonotes}
\usepackage{color, colortbl}
	
\definecolor{Gray}{gray}{0.9}
\definecolor{LightCyan}{rgb}{0.88,1,1}
\definecolor{LightBlue}{rgb}{236,244,255}
\usepackage{colortbl}
\usepackage{pgfplots}
\usepackage{hyperref}
\usepackage{appendix}
\AtBeginEnvironment{appendices}{\crefalias{section}{appendix}}
\BeforeBeginEnvironment{appendices}{\clearpage}

\usepackage{cleveref}
\crefname{section}{\S}{\S\S}
\crefname{table}{Tab.}{}
\crefname{figure}{Fig.}{Figs.}
\crefname{algorithm}{Algorithm}{}
\crefname{equation}{Eq.}{Eqs.}
\crefname{line}{Line}{}
\crefname{appendix}{App.}{}
\crefformat{section}{\S#2#1#3}
\crefname{thm}{Theorem}{}
\crefname{cor}{Corollary}{}
\crefname{prop}{Proposition}{}
\crefname{def}{Definition}{}

\usepackage{scalerel}
\usepackage[ruled,vlined]{algorithm2e}
\usepackage{linguex}
\usepackage{relsize}
\usepackage{array}
\usepackage{makecell}

\usepackage[font=small,labelfont=bf]{caption}
\usepackage{subcaption}
\usepackage{cancel}
\usepackage{inconsolata}

\usepackage{todonotes}

\definecolor{babyblue}{rgb}{0.54, 0.81, 0.94}

\definecolor{mygreen}{HTML}{7FC07F}
\definecolor{myred}{HTML}{DF7F7F}
\definecolor{mygray}{HTML}{BAB7B7}
\definecolor{pine}{HTML}{BCE672}
\definecolor{lotus}{HTML}{E4C6D0}
\definecolor{gold}{HTML}{F2BE45}
\definecolor{bamboo}{HTML}{789262}
\definecolor{jade}{HTML}{3DE1AD}
\definecolor{blueblue}{HTML}{4C8DAE}
\usepackage{stmaryrd}

\newcommand{\Verb}[1]{\textcolor{teal}{\textbf{\textit{#1}}}}

\newcommand{\redEnt}[1]{\textcolor{red}{\textbf{#1}}}
\newcommand{\bleuEnt}[1]{\textcolor{blue}{\textbf{#1}}}
\newcommand{\pinkEnt}[1]{\textcolor{magenta}{\textbf{#1}}}
\newcommand{\cyanEnt}[1]{\textcolor{cyan}{\textbf{#1}}}
\newcommand{\Omit}[1]{$\llbracket$\textbf{#1}$\rrbracket$}

\newcommand{\redEntOmit}[1]{\textcolor{red}{\Omit{#1}}}
\newcommand{\bleuEntOmit}[1]{\textcolor{blue}{\Omit{#1}}}
\newcommand{\pinkEntOmit}[1]{\textcolor{magenta}{\Omit{#1}}}
\newcommand{\cyanEntOmit}[1]{\textcolor{cyan}{\Omit{#1}}}
\newcommand{\grayEnt}[1]{\textcolor{gray}{\textbf{#1}}}
\newcommand{\greenEnt}[1]{\textcolor{mygreen}{\textbf{#1}}}
\newcommand{\orangeEnt}[1]{\textcolor{darkorange}{\textbf{\underline{#1}}}}

\newcommand{\Document}{\textsc{document}}
\newcommand{\Sentence}{\textsc{sentence}}

\newcommand{\Entity}{\textsc{entity}}
\newcommand{\Tense}{\textsc{tense}}
\newcommand{\Ellipsis}{\textsc{ellipsis}}
\newcommand{\Pronoun}{\textsc{pronoun}}
\newcommand{\ZeroPronoun}{\textsc{zeroPro}}
\newcommand{\Term}{\textsc{term}}
\newcommand{\Coref}{\textsc{coref}}
\newcommand{\Quotation}{\textsc{quote}}

\newcommand{\Ambiguity}{\textsc{ambiguity}}

\newcommand{\Masculine}{\textsc{masculine}}
\newcommand{\Feminine}{\textsc{feminine}}
\newcommand{\Neuter}{\textsc{neuter}}
\newcommand{\Epicene}{\textsc{epicene}}

\newcommand{\BWB}{$\mathcal{BWB}$}

\newcommand{\SRC}{\textsc{src}}
\newcommand{\REF}{\textsc{ref}}
\newcommand{\MT}{\textsc{mt}}

\newcommand{\SMT}{\textsc{smt}}
\newcommand{\OMTa}{\textsc{ggl}}
\newcommand{\OMTb}{\textsc{bd}}
\newcommand{\OMTc}{\textsc{bing}}
\newcommand{\MTS}{\textsc{sent}}
\newcommand{\MTD}{\textsc{doc}}
\newcommand{\HT}{\textsc{ht}}

\newcommand{\PE}{\textsc{pe}}
\newcommand{\bsc}[1]{\textbf{\textsc{#1}}}

\newcommand{\BLEU}{\textsc{bleu}}
\newcommand{\BlonD}{\textsc{blonde}}

\newcommand{\COMET}{\textsc{comet}}
\newcommand{\BERTScore}{\textsc{bert}}
\newcommand{\BLEURT}{\textsc{bleurt}}
\newcommand{\METEOR}{\textsc{meteor}}
\newcommand{\TER}{\textsc{ter}}

\newcommand{\fluency}{\textsc{fluency}}
\newcommand{\adequacy}{\textsc{adequacy}}
\newcommand{\testset}[1]{\textsc{part#1}}
\newcommand{\rater}[1]{\textsc{rater#1}}

\newcommand{\myFontSize}{\scriptsize}
\def\stripzero#1{\expandafter\stripzerohelp#1}
\def\stripzerohelp#1{\ifx 0#1\expandafter\stripzerohelp\else#1\fi}

\newcommand{\ZH}[1]{{\myFontSize \begin{CJK*}{UTF8}{gbsn} #1 \end{CJK*}}}
\newcommand{\ZHbig}[1]{{\begin{CJK*}{UTF8}{gbsn} #1 \end{CJK*}}}
\newcommand{\ethz}{\textsuperscript{$\zeta$}}
\newcommand{\msra}{\textsuperscript{$\gamma$}}

\title{Discourse Centric Evaluation of Machine Translation\\ with a Densely Annotated Parallel Corpus}

\author{ 
Yuchen Eleanor Jiang{\ethz}~\;~Tianyu Liu{\ethz}~\;~Shuming Ma{\msra}\\
\textbf{Dongdong Zhang{\msra}~\;~Mrinmaya Sachan{\ethz}~\;~Ryan Cotterell{\ethz}}\\
  $^{\ethz}$ETH Z\"{u}rich~\;~$^{\msra}$Microsoft Research Asia\\
    \texttt{\{\href{mailto:yuchen.jiang@inf.ethz.ch}{yuchen.jiang},\href{mailto:tianyu.liu@inf.ethz.ch}{tianyu.liu},\href{mailto:ryan.cotterell@inf.ethz.ch}{ryan.cotterell},\href{mailto:mrinmaya.sachan@inf.ethz.ch}{mrinmaya.sachan}\}@inf.ethz.ch }\\
\texttt{\{\href{mailto:shuming.ma@microsoft.com}{shuming.ma},\href{mailto:dongdong.zhang@microsoft.com}{dongdong.zhang}\}@microsoft.com}
}
\date{}
\begin{document}
\maketitle

\begin{abstract}


Several recent papers claim human parity at sentence-level Machine Translation (MT), especially in high-resource languages. 
Thus, in response, the MT community has, in part, shifted its focus to document-level translation. 
Translating documents requires a deeper understanding of the structure and meaning of text, which is often captured by various kinds of discourse phenomena such as consistency, coherence, and cohesion. 
However, this renders conventional sentence-level MT evaluation benchmarks inadequate for evaluating the performance of context-aware MT systems.
This paper presents a new dataset with rich discourse annotations, built upon the large-scale parallel corpus \BWB{} introduced in \citet{jiang-etal-2022-blonde}. 
The new \BWB{} annotation introduces four extra evaluation aspects, i.e., entity, terminology, coreference, and quotation, covering 15,095 entity mentions in both languages.
Using these annotations, we systematically investigate the similarities and differences between the discourse structures of source and target languages, and the challenges they pose to MT. 
We discover that MT outputs differ fundamentally from human translations in terms of their latent discourse structures. This gives us a new perspective on the challenges and opportunities in document-level MT.  
We make our resource publicly available to spur future research in document-level MT and the generalization to other language translation tasks.
\newline
\newline
\hspace{.5em}\includegraphics[width=1.25em,height=1.25em]{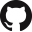}\hspace{.75em}\parbox{\dimexpr\linewidth-2\fboxsep-2\fboxrule}{\url{https://github.com/EleanorJiang/BlonDe/tree/main/BWB}}



\end{abstract}
\begin{figure*}[t]
\begin{adjustbox}{width=\textwidth}
\centering \small
\begin{tabular}{p{5pt}p{150pt}p{155pt}p{150pt}}
\toprule[2pt]
& SOURCE & REFERENCE & MT \\
\midrule[1pt]
1) & \ZH{	\redEnt{乔恋}攥紧了拳头，垂下了头。	}&	\redEnt{Qiao Lian} clenched \redEnt{her} fists and lowered \redEnt{her} head.	&	\redEnt{Joe} clenched \redEnt{his} fist and bowed \redEnt{his} head.	\\
2) & \ZH{	其实\bleuEnt{他}说得对。	}&	Actually, \bleuEnt{he} \Verb{was} right.	&	In fact, \bleuEnt{he}\Verb{'s} right.	\\
3) & \ZH{	\redEntOmit{}自己就是一个蠢货，竟然会[...]。	}&	\redEntOmit{She} was indeed an idiot, as only an idiot would [...]	&	\redEntOmit{I} \Verb{am} a fool, even \Verb{will} [...]	\\
5) & \ZH{	\redEnt{她}点进去，发现是\orangeEnt{凉粉群}，所有人都在@\redEnt{她}[...]	}&	\redEnt{She} logged into \redEntOmit{her} account and saw that a large number of fans in the \orangeEnt{Liang fan group} had tagged \redEnt{her}.[...]	&	\redEnt{She} nodded in and found it was a \orangeEnt{cold powder group}, and everyone was on \redEnt{her}.[...]	\\
7) & \ZH{【\greenEnt{川流不息}：\redEnt{乔恋}，快看\orangeEnt{微博}头条！ \orangeEnt{微博}头条！】}&	[\greenEnt{Chuan Forever}: \redEnt{Qiao Lian}, look at the headlines on \orangeEnt{Weibo}, quickly!]	&	\greenEnt{Chuan-flowing}: \redEnt{Joe love}, quickly look at the \orangeEnt{micro-blogging} headlines! \orangeEnt{Weibo} headlines?	\\
8) & \ZH{	\redEnt{她}微微一愣，拿起手机，登陆\orangeEnt{微博}，在看到头条的时候，整个人一下子愣住了！	}&	\redEnt{She} froze momentarily, then picked up \redEntOmit{her} cell phone and logged into \orangeEnt{Weibo}. When  \redEntOmit{she} saw the headlines,  \redEntOmit{her entire body} immediately froze over again!	&	\redEnt{She} took a slight look, picked up the phone, landed on the \orangeEnt{micro-blog}, when  \redEntOmit{she} saw the headlines,  \redEntOmit{the whole person} suddenly choked! \\
\bottomrule[2pt]
\end{tabular}
\end{adjustbox}
\caption{An example of the annotations in \BWB-test. The mentions in the same coreference chain are marked with the same color. 
Pronoun omissions are marked with \Omit{}.
The mistranslated verbs are marked with \Verb{teal},
and the mistranslated named entities are \orangeEnt{underlined}. 
(7) is a quotation with the speaker annotated as \greenEnt{Chuan Forever}.
The full chapter is in \Cref{fig:example_book1_0}. MT is the output of a Transformer-based sentence-level machine translation system. \looseness=-1}
\label{fig:intro_example}
\end{figure*}

\section{Introduction}
The field of machine translation (MT) has made tremendous strides in recent years thanks to neural machine translation (NMT) models that can utilize massive quantities of parallel training data~\citep[][\textit{inter alia}]{transformer, junczys-dowmunt-etal-2018-marian}. Unfortunately, most parallel corpora are aligned only at the sentence level, and document-level translation has remained limited to small-scale studies \citep{iwslt2020, opensubtitles2018, koehn-2005-europarl, tiedemann-2012-OPUS}.

There are, however, inherent differences between sentence-level translation and document-level translation, as documents consist of complex discourse structures that go beyond the mere concatenation of individual sentences. 
Three key discourse features are particularly important in document-level translation. First, the translations of \emph{named entities} have to be consistent. For example, in \Cref{fig:intro_example}, the same terminologies are not consistently referred to with the same translations (i.e. \orangeEnt{Weibo} vs. \orangeEnt{micro-blog}, \redEnt{Qiao Lian} vs. \redEnt{Joe} vs. \redEnt{Joe love}), and as a result the sentence-level MT system fails to capture discourse dependencies across sentences. 
Second, the \emph{coreference relationship} in the source language needs to be preserved. In particular, the relations between entities and their pronominal anaphora should be preserved, as well as the transition of discourse centers\footnote{Centers refer to the entities that are in the readers' focus at a certain point in the discourse. It can be realized by either pronouns or nominal mentions \cite{grosz-etal-1995-centering}.}. In \Cref{fig:intro_example} and \Cref{fig:example_book1_0}, coreference chains are color-coded, visually demonstrating the preservation of anaphoric referential relations and the transition chains of centers. 
Third, the \emph{conversational structure}, such as transitions between speakers, must be maintained.

Inferring latent discourse structures from documents is essential for translation coherence because of these discourse features. As a result, conventional sentence-level MT pipelines that do not consider context are unable to generate natural and coherent translations \citep{lapshinova-koltunski-etal-2018-parcorfull, werlen2021discourse}.
While efforts have been made to develop context-aware NMT models over the past few years~\citep{tiedemann-scherrer-2017-neural, agrawal2018contextual, voita-etal-2018-context, bawden-etal-2018-evaluating, ctx}, a reliable evaluation method for this type of MT has not yet been developed.

Traditional MT evaluation metrics, such as \BLEU~\cite{BLEU}, \textsc{ter}~\cite{TER}, \METEOR~\cite{Meteor}, are limited to evaluating translation quality at the sentence level and do not consider contextual information.
Even with the use of advanced neural evaluation methods such as \COMET~\cite{rei-etal-2020-comet}, \BLEURT~\cite{sellam-etal-2020-bleurt} and \textsc{BERTScore}~\cite{BERTScore}, it is not possible to accurately assess the coherence of translations within the context of a document.
\BlonD{}~\cite{jiang-etal-2022-blonde} has attempted to evaluate document-level translation quality and obtained a higher correlation with human evaluation. Yet, it is a coarse-grained automatic evaluation metric based on surface syntactic matching, thus being unable to evaluate deeper discourse features, such as coreferences and quotations.
\citet{vernikos2022embarrassingly} propose to concatenate context to extend pretrained metrics to the document level. However, its lack of interpretability leaves it unclear as to why it assigns a particular score to a given translation, which is a major reason why some MT researchers are reluctant to employ learned metrics in order to evaluate their MT systems~\cite{karpinska2022demetr}.

To address this issue, we have annotated a new benchmark based on the large-scale bilingual parallel corpus \BWB{} proposed in previous work~\cite{jiang-etal-2022-blonde}. 
This benchmark includes four evaluation aspects annotated on both source and target sides: entity, terminology, coreference, and quotation.
It consists of 80 documents across multiple fiction genres, 
with 15,095 mentions covered in 150,287 words in total. 

Using these annotations, we systematically evaluate the latent discourse structures of MT models.
We show that while context-aware MT systems perform better than phrase- and sentence-level ones in terms of entity translation, they still lag behind human translation in terms of proper noun and personal name translation.
In addition, we demonstrate that humans are far more adept at preserving coreference chains than MT models.

The main contributions of our paper are:
\begin{itemize} \setlength{\itemsep}{0pt} \setlength{\parskip}{0pt} \setlength{\parsep}{0pt}
    \item We propose a new benchmark that includes four types of discourse annotations that are closely related to document-level translation.
    \item We systematically investigate the similarities and differences between the discourse structures of source and target languages, and the challenges they pose to machine translation.
    \item We demonstrate that machine translations differ fundamentally from human translations in terms of their latent discourse structures. We believe these new insights can lead to potential improvements in future MT systems.
\end{itemize}

\section{Corpus}\label{sec:corpus}
\begin{figure}[t]
    \centering
    \includegraphics[width=0.3\textwidth]{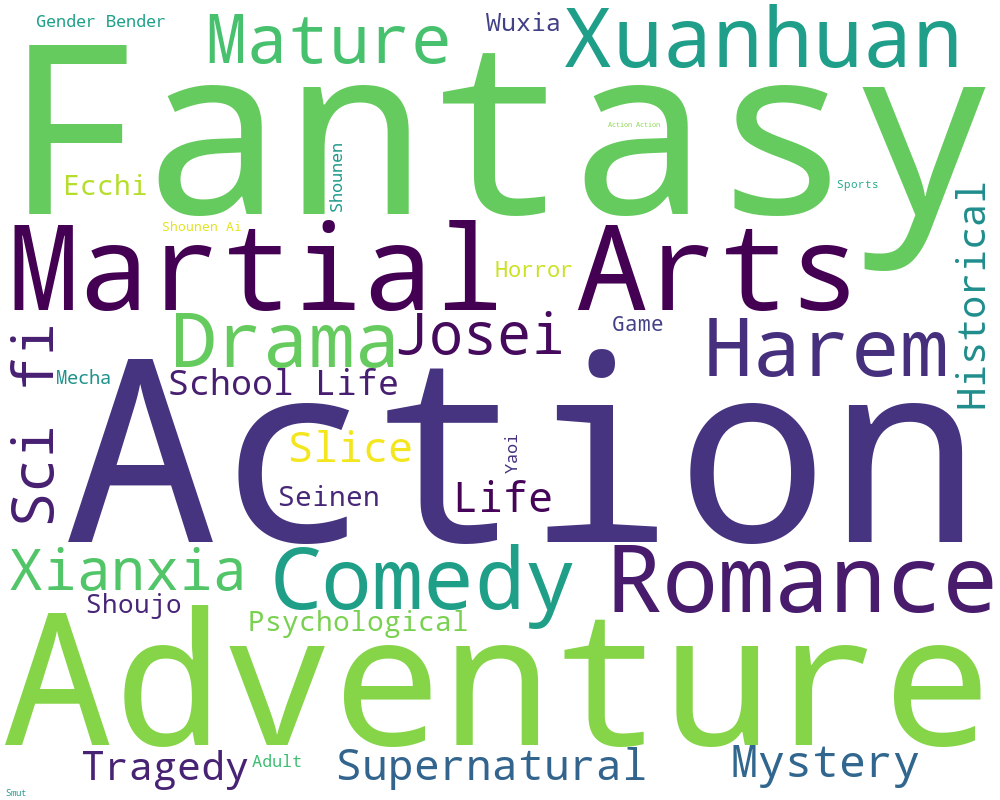}
    \caption{The genre distribution of novels in the \BWB{} corpus. Action, fantasy and adventure are the most common genres.\looseness=-1}
    \label{fig:genre_wordcloud}
\end{figure}

We draw our source material for annotation from the texts in the test set of \BWB{} \cite{jiang-etal-2022-blonde}, which consists of 65,107 tokens, 2,633 sentences in 80 different documents drawn from 6 web novels across different genres. 
The distribution of genres in \BWB{} are shown in \Cref{fig:genre_wordcloud}. 
The statistics of \BWB{} are shown in \Cref{tab:dataset_split}.
We refer the readers to \Cref{sec:datasetCreation} for the dataset creation details.


\begin{table}[htpb]
\centering
\begin{adjustbox}{width=\columnwidth}
\begin{tabular}{cccccccc}
\toprule[2pt]
   & \multicolumn{4}{c}{Size}  & \multicolumn{3}{c}{Averaged Length}  \\ \cmidrule(lr){2-5} \cmidrule(lr){6-8}
Split & \#word & \#sent & \#chap & \#book &  \#word/sent & \#sent/chap\\
\midrule[1pt]
Train & 325.4M & 9.57M                     & 196K                   & 378                        & 34.0                           & 48.8                            \\
Valid   & 68.0K  & 2.63K & 80 & 6                          & 25.8                           & 32.9                               \\
Test  & 67.4K  & 2.63K & 80 & 6                          & 25.7                           & 33.1                         \\
\bottomrule[2pt]
\end{tabular}
\end{adjustbox}
    \caption{Statistics of the \BWB{} corpus.}
    \label{tab:dataset_split}
\end{table}

\section{Annotation}\label{sec:annotation}
In this section, we describe the annotation scheme and the process of annotating the corpus, and provide examples of how the annotations can be used to study discourse phenomena in the two languages. 
The annotation was conducted by eight professional translators. 
A total of 8,585 mentions on the English side and 6,070 mentions on the Chinese side have been annotated.

\subsection{Named Entities} \label{subsec:entity}
The mistranslation of named entities (NEs) can significantly harm the quality of translation, although often under-reported by automatic evaluation metrics (e.g. \BLEU{}, \COMET{}). 
Therefore, we annotate named entities according to the ACE 2005 guidelines~\cite{ace05-entity}.  
These annotations identify six categories of entities: people (\texttt{PER}), facilities (\texttt{FAC}), geo-political entities (\texttt{GPE}), locations (\texttt{LOC}), vehicles (\texttt{VEH}), and organizations (\texttt{ORG}). In addition, each unique entity is assigned an \texttt{entity\_id} for the purpose of coreference resolution. For example, the following text includes annotations for 2 entities:

\exi.\label{ex:entity} $\cdots$ [\texttt{PER},6 the scion of [\texttt{ORG},5 the Ri family]]

Nested entities are also annotated. 
The annotated dataset contains a total of 5,984 English entities and 4,853 Chinese entities. 
The cause of the fewer entities in Chinese compared to English can be attributed to the subject omission phenomenon in Chinese discourse. This phenomenon also serves as a significant contributor to translation errors in context-agnostic MT systems when translating from Chinese to English, as shown in \Cref{fig:intro_example} (3).


\subsection{Terminology} \label{subsec:terminology}
Terminology refers to specialized words or phrases conventionally associated with a particular subject matter or whose use is agreed upon by a community of speakers.
In the context of a novel, terminology refers to the specific words and phrases that are used to describe the concepts, characters, settings, and events in the story. These words and phrases may be specific to the genre or style of the novel, or they may be unique to the world or setting created by the author. 
For example, in a fantasy novel that uses made-up magic terms, incorrect terminology translation could lead to misunderstandings of the powers and abilities of characters, the rules of the magical world, or the significance of certain events. Similarly, incorrect translations of named entities could lead to confusion or misunderstandings about the identities and roles of characters in the story. 
Inconsistent terminology translation can compromise the integrity and cohesion of a work and make it difficult for readers to fully grasp the intended meaning. 
Therefore, we have included a layer for terminology identification. This layer is a binary classification of whether a certain span counts as terminology (\texttt{T}) or not (\texttt{N}). 
\Cref{tab:entity_term} shows some term and non-term annotations in the dataset.
There are 2,156 English terms and 2,290 Chinese terms that have been annotated, accounting for approximately 52\% of all entities.

\begin{table}[]
    \centering
\begin{adjustbox}{width=\columnwidth}
\begin{tabular}{lll}
\textsc{source} & \textsc{reference} & \MT\\ \toprule
   $[_{\texttt{PER,T,1}}$ \ZHbig{木子} $]$ & $[_{\texttt{PER,T,1}}$ Mu Zi $]$ &  Wood\\
   $[_{\texttt{VEH,N,2}}$ \ZHbig{马车} $]$ & $[_{\texttt{VEH,N,2}}$ the carriage $]$ & the carriage\\
   $[_{\texttt{FAC,N,3}}$ \ZHbig{屋子} $]$ & $[_{\texttt{FAC,N,3}}$ the house $]$ &  the house\\
   $[_{\texttt{GPE,N,4}}$ \ZHbig{欧洲} $]$ & $[_{\texttt{GPE,N,4}}$ Europe $]$ &  Europe\\
   $[_{\texttt{LOC,N,7}}$ \ZHbig{林} $]$ & $[_{\texttt{LOC,N,7}}$ the forest $]$ &  the forest\\
   (omitted)  & $[_{\texttt{ORG,T,12}}$ the Ri family $]$ &  not translated \\
   $[_{\texttt{ORG,T,14}}$ \ZHbig{三泉宗} $]$ & $[_{\texttt{ORG,T,14}}$ the Three Spring Sect $]$ &  Sanquanzong\\
\end{tabular}
\end{adjustbox}
    \caption{Examples of entity annotations and terminology annotations. (1) Terminology is more prone to mistranslation and situations where translation is inconsistent with context. (2) MT often produces translations of poor quality when the Chinese subject is omitted. This is likely due to the difficulty in understanding the semantics of the sentence with ellipsis, as the subject is an integral part of the meaning of the sentence.}
    \label{tab:entity_term}
\end{table}

\subsection{Coreference} \label{subsec:coref}
Coreference is the phenomenon in which a word or phrase in a text refers to another word or phrase that has been previously mentioned in the text. 
Establishing coreference is important for determining gender/number marking on pronouns, determiners, and adjectives and for making lexical decisions (e.g. "I approached the town/river. I walked up to its bank.")
Errors can compromise coherence and accuracy. For example, consider the following sentence in English:
\\\\
\begin{adjustbox}{width=0.49\textwidth}
\begin{tabular}{ll}
\SRC &  In fairness, Miller did not attack \textcolor{myred}{the statue} itself. \\
& ... But he did attack \textcolor{myred}{its} meaning ... \\
\REF & Um fair zu bleiben, Miller griff nicht \textcolor{myred}{die Statue} selbst an. \\
& ... Aber er griff \textcolor{myred}{deren} Bedeutung an ... \\
\MT & Fairerweise hat Miller \textcolor{myred}{die Statue} nicht selbst angegriffen. \\
& ... Aber er griff \textcolor{myred}{seine} Bedeutung an ... \\
\end{tabular}
\end{adjustbox}
\\\\
In this example, both \textit{itself} and \textit{its} refer back to \textit{the statue}. However, in \MT{}, \textit{its} is incorrectly translated as \textcolor{myred}{seine}, which is the masculine possessive pronoun in German, as a result of the incorrect coreference resolution.

We follow the OntoNotes Coreference Annotation Guidelines for English and Chinese~\cite{ontonotes}, and consider proper noun phrases, common noun phrases, and personal pronouns as coreference candidates. 
In particular, we follow the three important principles of the OntoNotes guidelines:

\paragraph{Maximal Spans.} We include all modifiers in an annotated span, e.g. $[$ the surrounding passersby, who were actually reporters in disguise $]$.


\paragraph{Rather Lack Than Abuse.}  When in doubt, do not mark any coreferences.

\paragraph{Ellipsis.}
Omitted pronouns are marked with \texttt{O} and other pronouns are marked with \texttt{P}.
For instance, consider: 
\\\\
\begin{adjustbox}{width=0.49\textwidth}
\begin{tabular}{ll}
\SRC & $[_{\texttt{PER,T,1}}$\ZHbig{乔恋}$]$ \ZHbig{攥紧了拳头},$[_{\texttt{P,1}}$\ZHbig{她}$]$\ZHbig{垂下了头。} \\
\REF & $[_{\texttt{PER,T,1}}$Qiao Lian$]$ clenched $[_{\texttt{O,1}}]$ her$]$ fists and \\
& lowered $[_{\texttt{P,1}}$ her$]$ head. 
\end{tabular}
\end{adjustbox}
\\\\
In this example, the first \textit{her} is omitted in Chinese and is therefore marked as \texttt{<O, 1>}, where \texttt{1} is the entity id of \textit{Qiao Lian}.
\subsection{Quotations} \label{subsec:quotation}
The final annotation layer is quotation. 
In this stage of the process, we identify instances of direct speech and attribute the speech to its speaker. 
The inclusion of direct speech is common in literature, and its proper translation is essential. 
For instance, the same person can be addressed by different names by different people.
Furthermore, MT systems that lack contextual awareness may have difficulty correctly identifying the speaker in instances of direct speech, leading to inconsistencies in overall contextual translation.
For example, 

\exi. [{\texttt{Q,2}} ``Oh dear! Oh dear! I shall be late!'']

where \texttt{2} is the entity id of the speaker and \textit{``Oh dear!''} is an exclamation. In this example, discriminating exclamations from vocatives is vital for the cohesiveness of the story. 
In addition, there are cases where knowing the speaker is important for coreference. 
e.g., John/Mary said to Mary/John, "Oh, this is your dog." Her/His dog barked.
There are 840 (31.9\%) sentences that contain quotations, and there are 25 distinct speakers in total.

\paragraph{Inter-Annotator Agreement}
To ensure annotation quality, we randomly select 10 documents and have them independently annotated by another expert following previous work \cite{bamman-etal-2019-annotated}. We then calculate the mention overlap and F1 scores for entity, terminology, coreference, and quotation between the two annotations, using the regular annotator's work as the hypothesis and the second annotator's work as the reference. 
We achieve comparable inter-annotator agreements with previous work \cite{craft-coref,bamman-etal-2019-annotated}. The results are reported in \Cref{tab:iaa}.

\begin{table}[t]
\centering
\begin{adjustbox}{width=0.49\textwidth}
\begin{tabular}{cccccc}
\toprule[2pt]
 & Mention & Entity & Terminology & Coreference & Quotation\\
\midrule[1pt]
English  & 80.7 & 79.3 & 89.2 & 74.1 & 90.1\\
Chinese  & 86.5 & 74.1 & 84.8 & 65.9 & 89.7\\
\bottomrule[2pt]
\end{tabular}
\end{adjustbox}
    \caption{Inter-annotator agreement. For coreference, the average CoNLL F1 scores are reported.\looseness=-1}
    \label{tab:iaa}
\end{table}

\section{Bilingual Analysis} \label{sec:analysis}
We conduct a thorough bilingual analysis of the novel evaluation aspects using the new annotation.
The following two research questions are being investigated:
\begin{itemize}\setlength{\itemsep}{0pt} \setlength{\parskip}{0pt} \setlength{\parsep}{0pt}
    \item How different (or similar) are the discourse structures in the source language (Chinese) and the target language (English)?
    \item How do the differences in discourse structures affect the translation quality of MT systems?
    \vspace{-1ex}
\end{itemize}

\begin{table}[t]
\centering 
\begin{adjustbox}{width=0.49\textwidth}
\begin{tabular}{c|ccc|ccc}
\toprule[2pt]
 & \multicolumn{3}{c}{EN} & \multicolumn{3}{c}{ZH} \\
Type                 & Count     & Freq. & Repet.         & Count     & Freq. & Repet.     \\
\midrule[1pt]
\texttt{PER}  & 3,387 & 80.5\% & 16 & 3,552 & 81.6\% & 26 \\
\texttt{FAC}  & 360   & 8.6\%  & 2  & 325   & 7.5\%  & 3  \\
\texttt{ORG}  & 232   & 5.5\%  & 5  & 241   & 5.5\%  & 6  \\
\texttt{LOC}  & 108   & 2.6\%  & 3  & 115   & 2.6\%  & 4  \\
\texttt{VEH}  & 79    & 1.9\%  & 3  & 77    & 1.8\%  & 3  \\
\texttt{GPE}  & 44    & 1.0\%  & 1  & 41    & 0.9\%  & 1  \\
\midrule[1pt]
\texttt{NOM}  & 2,054 & 48.8\% & 8  & 2,061 & 47.4\% & 11 \\
\texttt{TERM} & 2,156 & 51.2\% & 18 & 2,290 & 52.6\% & 30 \\
\bottomrule[2pt]
\end{tabular}
\end{adjustbox}
    \caption{The distributions of different types of entities in both English and Chinese in \BWB{}-test. \textit{Freq.} and \textit{Repet.} stand for the frequency and the average number of repetitions, respectively.
    \looseness=-1}
    \label{tab:entity_type}
\end{table}
\begin{table}[t]
\centering 
\begin{adjustbox}{width=0.49\textwidth}
\begin{tabular}{cccccc}
\toprule[2pt]
Lang & \Masculine & \Feminine & \Neuter & \Epicene & \textbf{Omitted}\\
\midrule[1pt]
EN & 1,633 & 2,521 & 608 & 391 & \textbf{64.9\%} \\
ZH & 654 & 967 & 14 & 118 & 9.4 \% \\
\bottomrule[2pt]
\end{tabular}
\end{adjustbox}
    \caption{The distributions of different types of pronouns in both English and Chinese in the \BWB{} test set. 
    \looseness=-1}
    \label{tab:pronoun}
\end{table}
\paragraph{Entity types and terminology}
\Cref{tab:entity_type} demonstrates the distribution of entity annotations and term annotations in both English and Chinese in the \BWB{} test set.
First, \BWB{}-test is dominated by person and facility entities, with a much lower proportion of geo-political entities. In addition, terminology entities and person entities are repeated more frequently, requiring better translation consistency.
An important finding from our analysis is that the number and distribution of entities in Chinese and English are similar. This suggests that at the discourse level, the information being conveyed in these two languages is largely the same. 
This is consistent with the idea of ``language agnostic'' information, which implies the possibility of effectively transferring discourse-level knowledge between languages without needing to fully understand the specific language-specific nuances of each language in cross-lingual tasks~\citep{pan-etal-2017-cross, conneau-etal-2018-xnli}.


\paragraph{Pronouns}
Pronoun translation has been the focus of discourse-level MT evaluation~\cite{hardmeier-2012-discourse, APT-2017}.
We compare the numbers of different types of pronouns in Chinese and English in \Cref{tab:pronoun}.
As can be seen, Chinese has significantly fewer pronouns due to its pronoun-dropping property. This poses additional challenges for NMT systems, as they must be able to resolve anaphoric references.
In addition, \Cref{tab:pronoun} reveals a notable difference in the frequency of neuter pronouns. English exhibits a higher prevalence of neutral pronouns, possibly due to the presence of a larger number of expletive subjects in the language.

\paragraph{Coreference}
We then carry out several analyzes to investigate the differences and similarities between coreference behaviors in Chinese and English. 
\Cref{fig:coref_analysis}(a) examines the distribution in distances to the \emph{nearest} antecedent for both Chinese and English. The average distance between antecedents and anaphora in English is shorter than in Chinese. This may be connected to the more common use of pronoun ellipsis in Chinese --- pronouns are often omitted when referring to closer antecedents. 
\Cref{fig:coref_analysis}(b) illustrates the distance between the first and last mention of an entity within each coreference chain. It can be observed that, although English coreference chains tend to be longer in general compared to Chinese coreference chains, the distribution of these lengths is consistent. This represents another language-independent discourse feature in addition to entity distribution. The spread distributions of the two languages are further depicted in \Cref{fig:example_book1_0} and \Cref{fig:example_book153_0}.
Finally, we present an analysis of the size of coreference chains in \Cref{fig:coref_analysis}(c). Our results indicate that the number of mentions in English coreference chains tends to be larger than in Chinese.

\newcommand{\setfigheight}{18pt}       
\newcommand{\imgtrimtopsize}{1cm} 
\newcommand{\imgtrimleftsize}{1cm}
\newcommand{\imgtrimrightsize}{1cm}
\newcommand{\imgtrimbottomsize}{2cm}  
\newcommand{\setfigwidth}{0.33\textwidth}   


\begin{figure*}[t]
    \centering
    \begin{adjustbox}{width=\linewidth}
    \begin{tabular}{ccc}
    \includegraphics[width=\setfigwidth,trim={\imgtrimleftsize{} 0 5cm 0},clip]{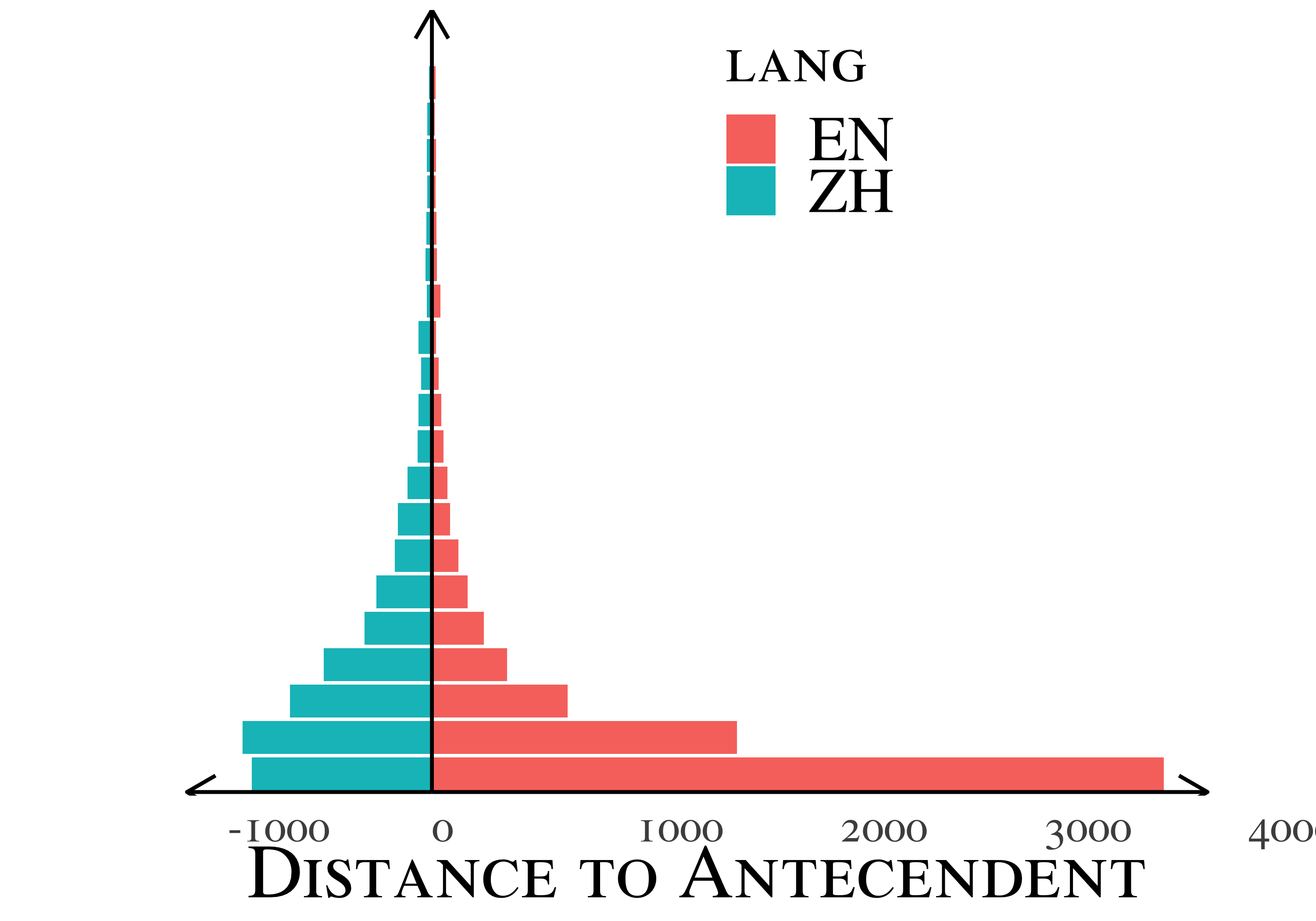} &
    \includegraphics[width=\setfigwidth,trim={\imgtrimleftsize{} 0 0 0},clip]{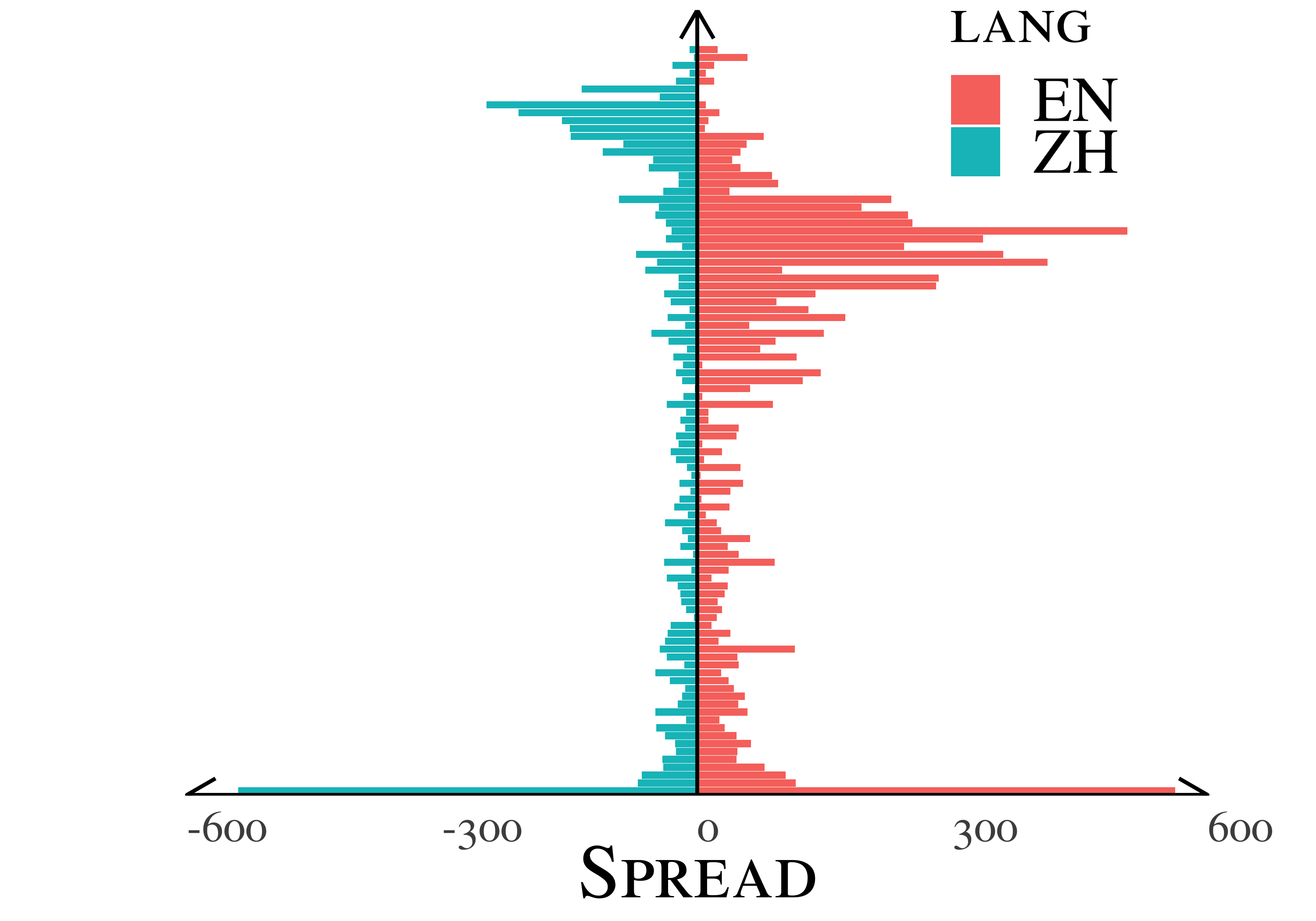} &  
     \includegraphics[width=\setfigwidth,trim={\imgtrimleftsize{} 0 0 0},clip]{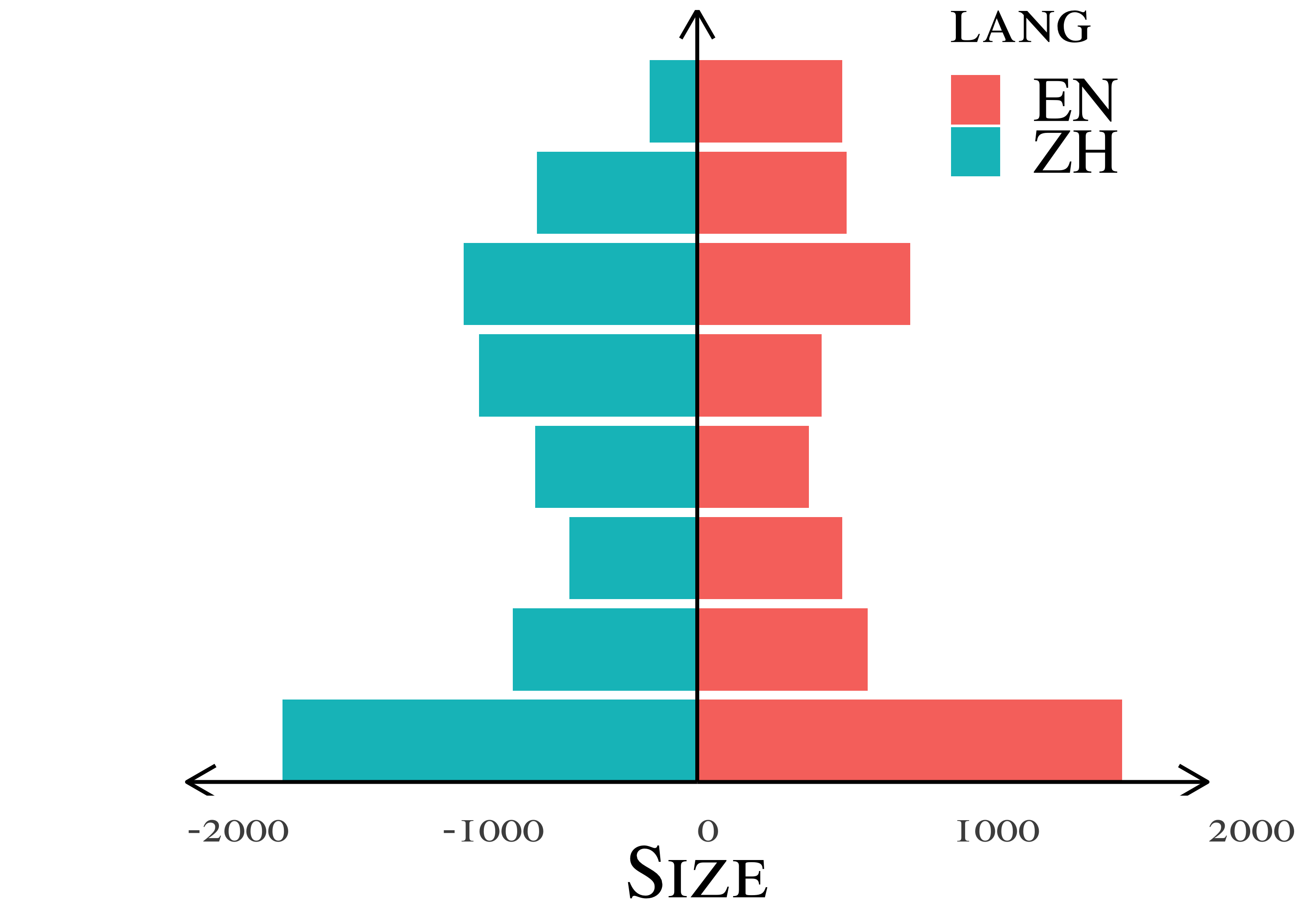} 
    \end{tabular}
    \end{adjustbox}
    \caption{The histograms of coreference chains. The blue and red bars are the count bins of the distance to the nearest antecedent for each mention, the spreads between the first and last mention of entity, and the number of mentions in each coreference chain, respectively. \looseness=-1}
    \label{fig:coref_analysis}
\end{figure*}

\begin{table}[t]
\centering
\begin{adjustbox}{width=0.49\textwidth}
\begin{tabular}{ccp{120pt}c}
\toprule[2pt]
Error Type & \# & Description & An\\
\midrule[1pt]
\Entity  & 43.3\% & error(s) due to the mistranslation of named entities. & \CheckmarkBold \\
\Term  & 19.2\% & error(s) caused by the mistranslation of terminologies. & \CheckmarkBold \\
\Coref  & 34.0\% & error(s) caused by coreference resolution failure(s). & \CheckmarkBold \\
\ZeroPronoun  & 17.3\% & error(s) caused by the omission of pronoun(s). & \CheckmarkBold \\
\Quotation  & 1.1\% & error(s) caused by the misinterpretation of quotation(s). & \CheckmarkBold \\
\bottomrule[2pt]
\end{tabular}
\end{adjustbox}
    \caption{The types of NMT errors and their description. \# represents the percentage of the error in the \BWB{} test set. \CheckmarkBold indicates ``with annotation''. \looseness=-1}
    \label{tab:errors}
\end{table}

\paragraph{Challenges for MT} \label{subsec:error}
So far our analyses have revealed that the Chinese source language and English target language exhibit several distinct linguistic characteristics, leading to various discourse phenomena, including the omission of pronouns in Chinese and the use of dummy subjects in English, as well as shorter English coreference chains, among others. These discourse phenomena present challenges for MT. In order to systematically analyze translation errors caused by these phenomena, we conducted a study in which four professional translation experts compared reference translations to those generated by a commercial MT system Google Translate.
The annotation guidelines are as follows:
\begin{enumerate} \setlength{\itemsep}{0pt} \setlength{\parskip}{0pt} \setlength{\parsep}{0pt}
    \vspace{-0ex}
    \item Identify translation errors and identify whether the translation error is at the \Document{} level, e.g. the translation is inconsistent with the context or does not comply with the global criterion of coherence.
    \item Categorize the \Document{} errors in accordance with the discourse phenomena and mark the corresponding spans in the reference (English) that cause the MT output to be incorrect.
    \vspace{-3ex}
\end{enumerate}
The results, summarized in \Cref{tab:errors}, indicate that named entity translation is the most significant issue, with terminology translation, coreference resolution failures, and pronoun omission also being notable problems. This analysis emphasizes the importance of accurately inferring latent discourse structures in the translation of long texts.

\begin{table*}[t]
\centering 
\begin{adjustbox}{width=\textwidth} 
\begin{tabular}{c|p{1cm}<{\centering}p{1cm}<{\centering}p{1cm}<{\centering}p{1cm}<{\centering}p{1cm}<{\centering}p{1cm}<{\centering}|cc|cc|cc|cc|cc}
\toprule[2pt]
 & \multicolumn{6}{c|}{Automatic Metrics} & \multicolumn{10}{c}{Discourse Phenomena} \\
& \BLEU & \METEOR & \TER &\BERTScore &\COMET & \BlonD & \multicolumn{2}{c|}{\Ambiguity} & \multicolumn{2}{c|}{\Entity} & \multicolumn{2}{c|}{\Tense} & \multicolumn{2}{c|}{\Pronoun} & \multicolumn{2}{c}{\Ellipsis} \\
\midrule[1pt]
\SMT  & \cellcolor[HTML]{FDF8F8}6.86  & \cellcolor[HTML]{FEFEFE}18.73 & \cellcolor[HTML]{73A7D6}84.35 & \cellcolor[HTML]{FFFFFF}32.14 & \cellcolor[HTML]{FFFFFF}-0.65 & \cellcolor[HTML]{FEFEFE}15.52 & \cellcolor[HTML]{B4D0E9}28.92 & \cellcolor[HTML]{C1D4E9}20.94 & \cellcolor[HTML]{D4E4F2}30.05 & \cellcolor[HTML]{E0EBF6}19.67 & \cellcolor[HTML]{FEFEFE}51.01 & \cellcolor[HTML]{FFFFFF}44.58 & \cellcolor[HTML]{FFFFFF}52.21 & \cellcolor[HTML]{FFFFFF}38.73 & \cellcolor[HTML]{FFFFFF}34.85 & \cellcolor[HTML]{FFFFFF}21.66 \\
\OMTb & \cellcolor[HTML]{CEDCEC}10.02 & \cellcolor[HTML]{3D85C6}26.57 & \cellcolor[HTML]{FEFEFE}70.72 & \cellcolor[HTML]{86B3DC}46.78 & \cellcolor[HTML]{82B1DB}-0.16 & \cellcolor[HTML]{F9FBFD}16.17 & \cellcolor[HTML]{87B3DC}41.52 & \cellcolor[HTML]{9DBEDF}31.47 & \cellcolor[HTML]{FEFEFE}21.28 & \cellcolor[HTML]{FEFEFE}12.04 & \cellcolor[HTML]{BAD3EB}58.36 & \cellcolor[HTML]{AFCDE8}54.19 & \cellcolor[HTML]{D7E6F4}56.81 & \cellcolor[HTML]{E3EDF7}43.65 & \cellcolor[HTML]{AFCDE8}52.53 & \cellcolor[HTML]{BED6EC}41.05 \\
\OMTc & \cellcolor[HTML]{ACC8E3}12.23 & \cellcolor[HTML]{8FB8DE}23.27 & \cellcolor[HTML]{3D85C6}89.53 & \cellcolor[HTML]{B2CFE9}41.45 & \cellcolor[HTML]{BAD3EB}-0.38 & \cellcolor[HTML]{C3D9ED}22.47 & \cellcolor[HTML]{FEFEFE}8.09  & \cellcolor[HTML]{FDF8F8}3.33  & \cellcolor[HTML]{CDE0F0}31.37 & \cellcolor[HTML]{D4E4F2}22.62 & \cellcolor[HTML]{B7D2EA}58.66 & \cellcolor[HTML]{A9C9E6}54.93 & \cellcolor[HTML]{ADCCE7}61.54 & \cellcolor[HTML]{B0CEE8}52.29 & \cellcolor[HTML]{FFFFFF}34.88 & \cellcolor[HTML]{FFFFFF}21.74 \\
\OMTa & \cellcolor[HTML]{A3C2E1}12.81 & \cellcolor[HTML]{CCDFF0}20.80 & \cellcolor[HTML]{91BADF}81.39 & \cellcolor[HTML]{86B3DC}46.80 & \cellcolor[HTML]{92BBDF}-0.22 & \cellcolor[HTML]{C0D7EC}22.86 & \cellcolor[HTML]{74A8D6}46.66 & \cellcolor[HTML]{8DB5DB}36.05 & \cellcolor[HTML]{E0ECF6}27.46 & \cellcolor[HTML]{DFEBF5}20.03 & \cellcolor[HTML]{A1C4E3}61.01 & \cellcolor[HTML]{8BB6DD}58.58 & \cellcolor[HTML]{C3DAEE}59.04 & \cellcolor[HTML]{C6DBEF}48.63 & \cellcolor[HTML]{DAE8F5}43.04 & \cellcolor[HTML]{E1ECF6}30.76 \\
\MTS  & \cellcolor[HTML]{7EACD7}15.24 & \cellcolor[HTML]{A4C6E4}22.39 & \cellcolor[HTML]{AACAE6}78.93 & \cellcolor[HTML]{B1CEE8}41.58 & \cellcolor[HTML]{6DA3D4}-0.08 & \cellcolor[HTML]{9FC2E3}26.79 & \cellcolor[HTML]{4B8ECA}58.16 & \cellcolor[HTML]{88B2DA}37.59 & \cellcolor[HTML]{FAFBFD}22.27 & \cellcolor[HTML]{DDE9F5}20.54 & \cellcolor[HTML]{4F91CC}69.77 & \cellcolor[HTML]{4E90CB}65.77 & \cellcolor[HTML]{8DB7DE}65.21 & \cellcolor[HTML]{74A8D6}62.64 & \cellcolor[HTML]{7BACD8}64.01 & \cellcolor[HTML]{8BB7DD}55.81 \\
\MTD  & \cellcolor[HTML]{5D98CF}17.45 & \cellcolor[HTML]{5F9AD0}25.21 & \cellcolor[HTML]{3E86C7}89.48 & \cellcolor[HTML]{97BDE1}44.78 & \cellcolor[HTML]{5091CC}0.03  & \cellcolor[HTML]{76A9D7}31.53 & \cellcolor[HTML]{3D85C6}61.95 & \cellcolor[HTML]{74A6D5}43.18 & \cellcolor[HTML]{DDE9F5}28.25 & \cellcolor[HTML]{C8DCEF}25.67 & \cellcolor[HTML]{4A8DCA}70.39 & \cellcolor[HTML]{3D85C6}67.80 & \cellcolor[HTML]{3D85C6}74.20 & \cellcolor[HTML]{3D85C6}71.96 & \cellcolor[HTML]{4F91CC}73.55 & \cellcolor[HTML]{5594CD}71.93 \\
\PE   & \cellcolor[HTML]{3D85C6}19.52 & \cellcolor[HTML]{A5C6E4}22.38 & \cellcolor[HTML]{AFCDE7}78.47 & \cellcolor[HTML]{3D85C6}55.50 & \cellcolor[HTML]{3D85C6}0.10  & \cellcolor[HTML]{3D85C6}38.18 & \cellcolor[HTML]{4288C8}60.65 & \cellcolor[HTML]{3D85C6}59.09 & \cellcolor[HTML]{3D85C6}60.94 & \cellcolor[HTML]{3D85C6}60.41 & \cellcolor[HTML]{3D85C6}71.68 & \cellcolor[HTML]{4087C7}67.56 & \cellcolor[HTML]{5594CD}71.51 & \cellcolor[HTML]{7DAED9}61.04 & \cellcolor[HTML]{3D85C6}77.45 & \cellcolor[HTML]{3D85C6}78.75 \\
\bottomrule[2pt]
\end{tabular}
\end{adjustbox}
\caption{Results of MT systems and human post-editing on the \BWB{} test set. For discourse phenomena, we report both F1 measure defined in \citet{jiang-etal-2022-blonde} and Exact-Match Accuracy defined in \citet{alam-terminology-2021}.  \looseness=-1}
\label{tab:scores}
\end{table*} 
\begin{figure}[t]
    \centering
    \includegraphics[width=0.45\textwidth,trim={0cm 7.5cm 1.5cm 0.5cm} ,clip]{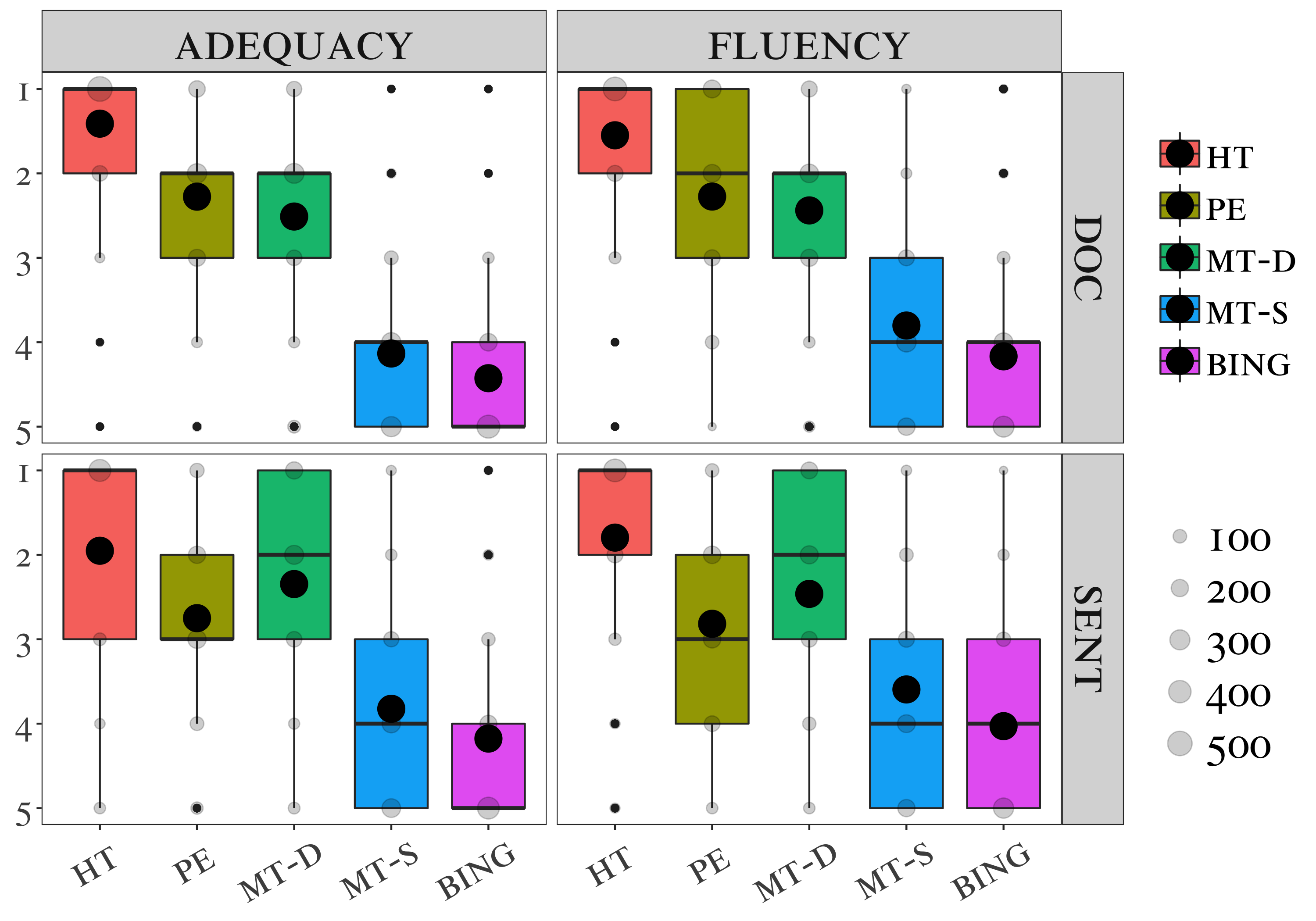}
    \caption{Human evaluation results on \BWB{}. Each $\bullet$ represents the average of the system's rankings. \looseness=-1}
    \label{fig:human_evaluation}
\end{figure}
 
\section{Exploring Discourse Features of MT Systems and Human Translation} \label{sec:exp}
The existence of a bilingual corpus with discourse annotations allows us to test the performance of both human translations and MT systems -- including the popularly used commercial systems and those trained on the in-domain dataset \BWB{}.
The following 6 MT systems are adapted:\footnote{\MTS{} and \MTD{} are trained on \BWB{} by fairseq~\cite{fairseq}, and the training details are in App.\ref{app:model_parameters}.}
\begin{itemize} \setlength{\itemsep}{0pt} \setlength{\parskip}{0pt} \setlength{\parsep}{0pt}
    \item \bsc{\SMT{}}: phrase-based baseline~\cite{SMT}.
    \item \bsc{\OMTc}, \bsc{g}oo\bsc{gl}e, \bsc{b}ai\bsc{d}u: commercial systems.
    \item \bsc{\MTS}: the Transformer baseline that translates sentence by sentence \cite{transformer}.
    \item \bsc{\MTD}: the document-level NMT model that adopts two-stage training \cite{ctx}.
    \item \bsc{\PE}: the post-edited \OMTc{} outputs produced by professional translators. They were instructed to correct only discourse-level errors with minimal modification. 

\end{itemize}

\paragraph{Overall Quality}
The quality of the translations produced by each system was first evaluated using a range of automatic translation metrics, including \BLEU, \METEOR, \BERTScore, \COMET, and \BlonD. The results of this evaluation are presented in \Cref{tab:scores}. In addition, we conducted a human evaluation of the translations, the results of which are shown in \Cref{fig:human_evaluation}.\footnote{App. \ref{app:human_evaluation} describes how human assessment is carried out. The inter-rater agreement is reported in \Cref{tab:kappa}.}
The large gap between the performances of \HT{} and \MT{} indicates that the genre of \BWB{}, i.e., literary translation, is challenging for \MT{}, and NMT systems are far beneath human parity.
\MTD{} performs significantly better than \MTS{}, suggesting that \BWB{} contains rich discourse phenomena that can only be translated accurately when the context is taken into account.
It is also worth noting that even though \PE{} is the post-edit of the relatively poor-performing system \OMTc{}, it still achieves surprisingly better performance than \MTD{} at the document level.
This observation confirms our claim that discourse phenomena have a huge impact on human judgment of translation quality.

\begin{figure}[t]
    \centering
    \includegraphics[width=0.4\textwidth,trim={0cm 0cm 0cm 0.5cm} ,clip]{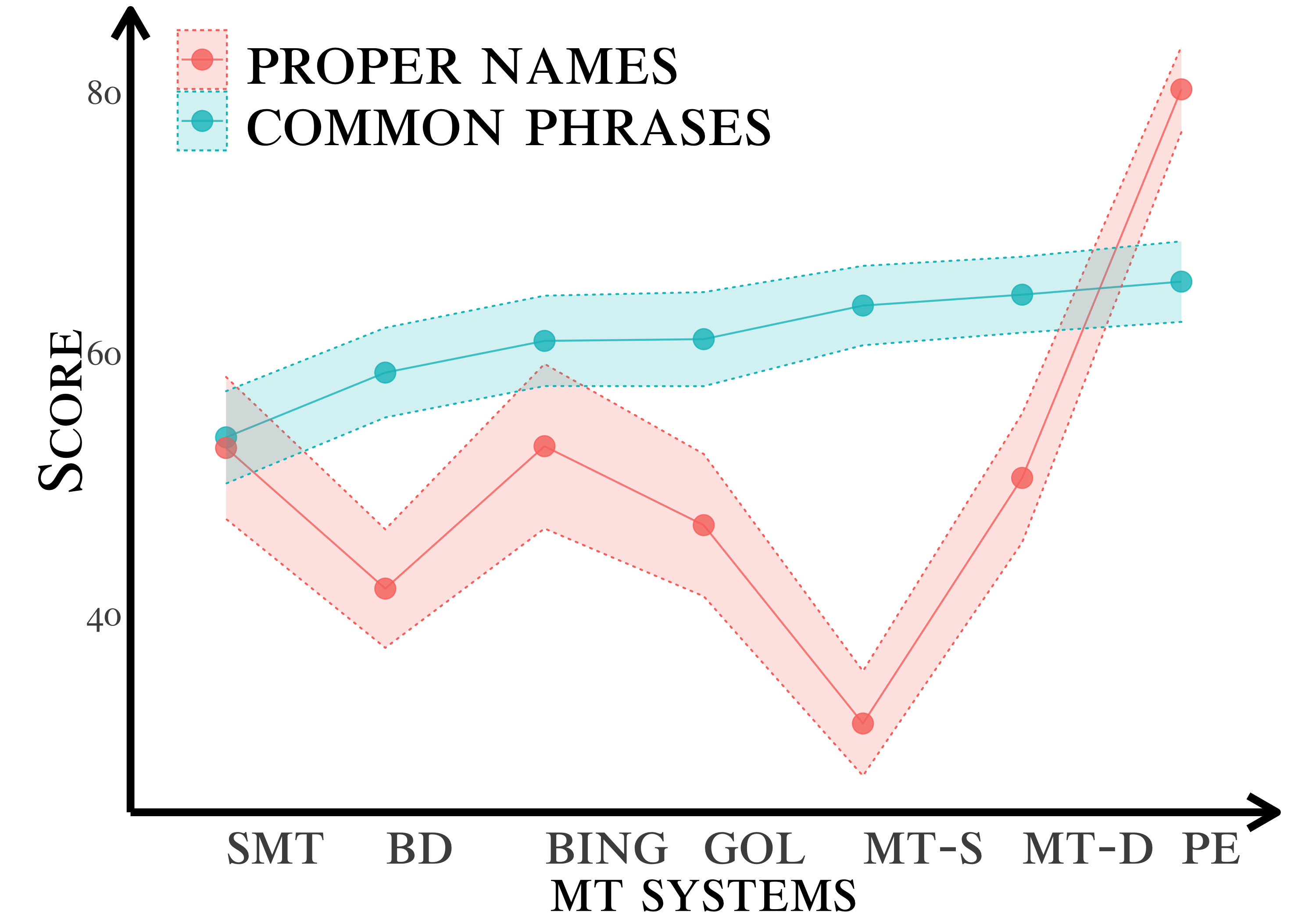}
    \caption{Terminology evaluation results on \BWB{}\looseness=-1}
    \label{fig:term_evaluation}
\end{figure}

\paragraph{Terminology Translation}
We next turn to evaluating the performance of each system on terminology translation, depicted in \Cref{fig:term_evaluation}. The recall rate of terminology is reported. As can be seen, although the accuracy of each system in the translation of common phrases is not significantly different, the performance of the \PE{} system is superior to that of the MT systems in the translation of proper names. Notably, despite having been trained on in-domain data, the \MTS{} system is unable to accurately translate terms on the test set due to the varied terminologies used in different novels, resulting in a performance that is even lower than that of commercial MT systems.

\begin{figure}[t]
    \centering
    \includegraphics[width=0.4\textwidth,trim={0cm 0cm 0cm 0.5cm} ,clip]{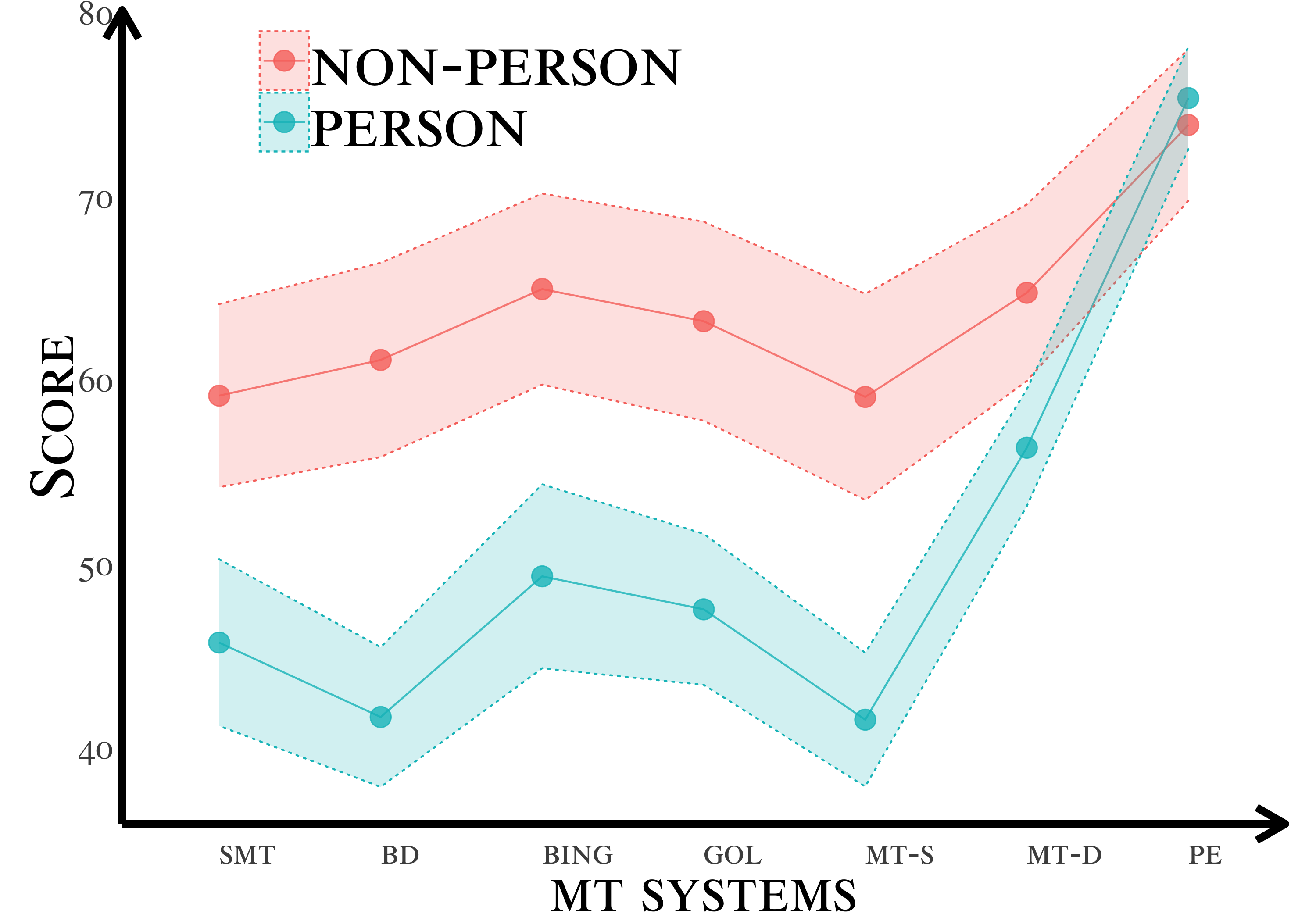}
    \caption{The F1 scores of person entities vs non-person entities on \BWB{}.\looseness=-1}
    \label{fig:per_evaluation}
\end{figure}
\begin{figure}[t]
    \centering
    \includegraphics[width=0.49\textwidth,trim={3cm 0cm 0cm 0cm} ,clip]{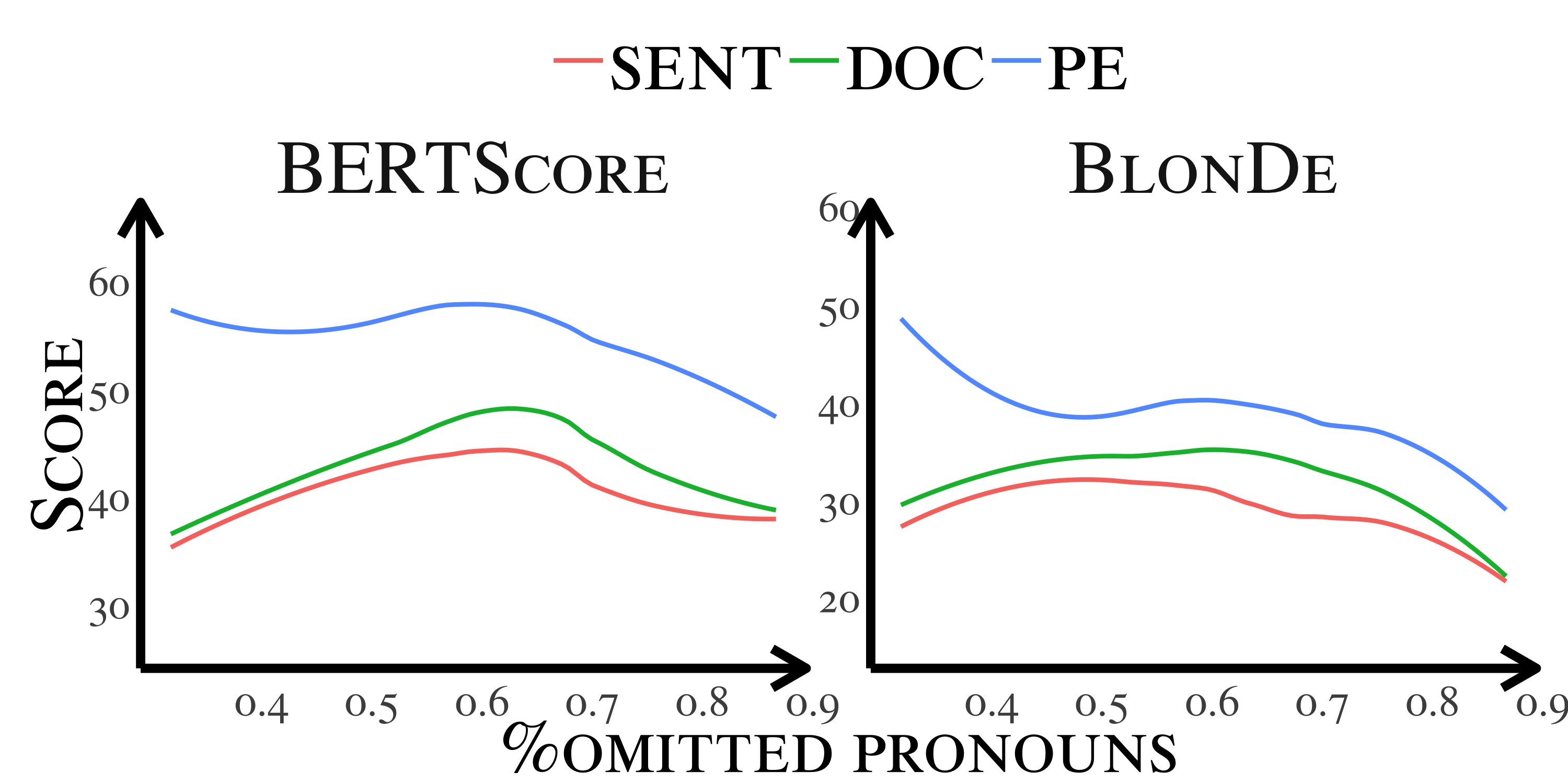}
    \caption{The \BlonD{} and \BERTScore{} as a function of omitted pronouns. The x-axis is the ratio of omitted pronouns to the total number of pronouns in each document.\looseness=-1}
    \label{fig:pro_evaluation}
\end{figure}

\paragraph{Entity Translation}
For the evaluation of entity translations, we compared translations of \texttt{PER} and translations of other entities. As illustrated in \Cref{fig:per_evaluation}, the translation of personal names and terms exhibits a similar trend, with the MT systems performing significantly worse than \PE{}. This outcome is expected, as both personal names and terms present similar challenges for MT, including (1) the need for consistent translation, (2) the potential for multi-word personal names to be combined in specific ways that may not translate well when each word is translated individually, and (3) the possibility of personal names not having a direct equivalent in other languages. For example, in \Cref{fig:example_book153_0}, the character name \redEnt{Ye Qing Luo} \textit{(literal: night, clear, fall)} is mistranslated into \redEnt{The night fell}.
Additionally, wordplay and puns based on character names may be difficult to convey in translation, as character names may be chosen for their sound or meaning in the original language, which may not translate effectively into other languages.

\paragraph{Pronoun Translation}
Our evaluation of pronoun translation focuses on the impact of omitted pronouns on translation quality. As depicted in \Cref{fig:pro_evaluation}, an increase in the proportion of omitted pronouns leads to a decrease in the translation quality of the human post-editing system. This finding suggests that pronoun omission is a major issue in Chinese-to-English translation. However, the \BERTScore{} scores of the sentence-level and document-level MT systems do not follow a monotonic trend with the increase in the proportion of omitted pronouns. Rather, the highest value is reached when the proportion of omitted pronouns is within the interval of $0.6-0.7$. This result indicates that NMT models process pronoun omissions in a manner that is fundamentally different from that of humans. In contrast, the \BlonD{} scores exhibit a pattern that is consistent with the trend observed in the \PE{} system, indicating that while the omission of pronouns may not significantly affect the quality of translation at the sentence level, it is crucial for ensuring the overall quality of document-level translation.

\begin{table}[t]
\centering
\begin{tabular}{cccc}
\toprule[2pt]
 & $B^3$\textuparrow & $CEAF$\textuparrow & \textsc{coherence}\textdownarrow\\
\midrule[1pt]
\MTS  & 43.3 &  49.7 & $\mathfrak{4}\,$ \\
\MTD  & 63.5 & 55.1 & $\mathfrak{3}^*$ \\
\PE  & 74.0 &  67.3 &  $\mathfrak{2}\,$ \\
\HT  & 88.1 & 76.7 & $\mathfrak{1}^*$ \\
\bottomrule[2pt]
\end{tabular}
    \caption{The coreference resolution performance ($B^3$, $CEAF$) and the coherence ranking of different translation systems. The Kruskal-Wallis test was applied and the systems that are significantly better than the previous one with p-value $> .05$ are marked with $^*$.\looseness=-1}
    \label{tab:coref}
\end{table}

\paragraph{Translation Coherence}
Finally, the coreference annotations allow us to measure the coherence of translated texts.
We first fine-tune a neural coreference model~\cite{lee-etal-2017-end} trained on OntoNotes~\cite{weischedel-2013-ontonotes}. This end-to-end model jointly performs mention detection and antecedent linking. 
We modify the model by replacing the span representations with SpanBERT embeddings~\cite{joshi-etal-2020-spanbert}. Additionally, during inference, we inject mentions that match the reference annotations into the antecedent candidate sets after the mention detection stage. The model then greedily selects the most likely antecedent for each mention in a document, or a dummy antecedent that begins a new coreference chain with the mention.
Coreference resolution performance and the coherence ranking are reported in \Cref{tab:coref}. Coherence is compared at the document level according to the centering theory~\cite{grosz-etal-1995-centering}. We use the operationalization of \citet{jiang2022investigating} and compare \textsc{kp} scores of each document.

As shown in \Cref{tab:coref}, \HT{} surpasses \MT{} in terms of entity coherence. Additionally, human post-editing was found to achieve entity coherence comparable to that of \HT{}. These results highlight the challenges that MT currently faces in regard to maintaining contextual coherence when translating longer texts.


\section{Comparing to Related Work} \label{sec:related_work}
\paragraph{Evaluation Test Suites for Context-Aware MT}

Several context-aware test suites have been proposed in recent years~\citep{DiscoMT, PROTEST, burchardt-etal-2017-linguistic, isabelle-etal-2017-challenge, rios-gonzales-etal-2017-improving, muller-etal-2018-large,bawden-etal-2018-evaluating,voita-etal-2019-good, guillou-hardmeier-2018-automatic}.
Although they facilitate the development of context-aware machine translation, they are not without limitations: First, the previous test suites are limited in size, with the largest being Parcorfull~\cite{lapshinova-koltunski-etal-2018-parcorfull} having 82,379 tokens from three different domains. Moreover, it is often the case that test sets are annotated based on a small portion of a parallel corpus that comprises documents. However, the parallel corpus from which previous test sets are selected is generally quite small, which makes it impossible to train large in-domain NMT models. In contrast, our test set is based on the ultra-large-scale parallel corpus \BWB, which is twice as large as the OpenSubtitles fr-en corpus~\cite{opensubtitles2018} upon which most previous challenge sets were based (see \Cref{tab:datasets_full}). This enables us to differentiate between the source of mistranslation being underfitting or the model not having the capability to model discourse structure. 
In addition, the scope of most test sets has been restricted to a single discourse phenomenon (the majority of which focus on pronoun translation), which makes it impossible to objectively compare a model's translation capability for different discourse phenomena in the same language and domain. For example, it is hard to decipher whether the model misinterpreted a pronoun's gender as a result of a coreference resolution error or as a consequence of misjudging the gender of the entity. In sharp contrast to this, we have four annotation layers on the same parallel corpus.

\paragraph{Monolingual Corpora with Discourse Annotations}

Linguistically annotated corpora have contributed significantly to the advancement of key natural language technologies such as named entity recognition \citep{tjong-kim-sang-de-meulder-2003-introduction}, coreference resolution \cite{lee-etal-2017-end}, and discourse parsing \cite{surdeanu-etal-2015-two}.
The majority of evaluation has, however, only been conducted on monolingual corpora such as the BBN name entity and pronoun coreference corpus~\cite{weischedel2005bbn}, the Penn Discourse Treebank~\citep{miltsakaki2004penn, webber2019penn}, and the OntoNotes corpus~\cite{weischedel-2013-ontonotes}.
And yet, languages differ considerably in terms of the discourse phenomena they exhibit. 
In particular, different languages have different linguistic features that influence the application of cohesive devices and there are language-specific constraints governing the choice of referring expressions.
For instance, \Cref{fig:intro_example} demonstrates a prominent feature that distinguishes Chinese from English. In Chinese, it is a common practice  to omit pronouns (referred to as pronoun-dropping phenomenon), and an anaphoric link can be inferred from context without explicit assertion. English, in contrast, does not allow the pronouns to be omitted. 
Our dataset contains aligned parallel discourse annotations in two languages, allowing analysis of the transferability of current NER and coreference resolution models across languages.




\section{Conclusion}
In this paper, we introduced the newly constructed bilingual parallel corpus \BWB{}-test, which includes annotations of various discourse phenomena. Our analysis of these annotations revealed the significant challenges posed by discourse phenomena to MT. Therefore, we advocate for greater attention on discourse coherence and consistency of the outputs of NMT models. The main discourse challenges faced by MT include entity consistency, entity recognition, anaphoric information loss, and coreference. Additionally, the \BWB{} corpus, with its rich discourse annotations, serves as a valuable resource for a variety of purposes, including studying the transferability of named entity recognition and coreference resolution, as well as the development of multilingual structured prediction models.

\section*{Limitations}
There are several limitations to the current study. Firstly, as for now, the corpus used in this study only consists of a single language pair. Secondly, the coherence of the MT systems was evaluated using a fine-tuned conference model, as no annotations were available for the MT outputs. However, as shown in \Cref{tab:coref}, the fine-tuned conference model is not perfect and may affect the quality of our coherence evaluation. Thirdly, this paper focuses on using discourse annotations to reveal and analyze discourse phenomena and the challenges they present to machine translation. Using the annotations to improve MT models is beyond the scope of this study and is left for future work.

\section*{Ethical Considerations}
The annotators were paid a fair wage and the annotation process did not solicit any sensitive information from the annotators. 
In regard to the copyright of our dataset, as stated in the paper, the crawling script that we plan to release will allow others to reproduce our dataset faithfully and will not be in breach of any copyright. 
In addition, the release of our annotated test set will not violate the doctrine of \textbf{Fair Use} (US/EU), as the purpose and character of the use is \emph{transformative}.
Please refer to \url{https://www.nolo.com/legal-encyclopedia/fair-use-the-four-factors.html} for relevant laws.


\bibliography{mine}
\bibliographystyle{acl_natbib.bst}

\clearpage
\begin{appendices}
\section{Document-Level Parallel Corpora} \label{app:dataset}
There are some document-level parallel corpora in the market: TED Talks of IWSLT dataset~\cite{iwslt2020}, News Commentary~\cite{tiedemann-2012-OPUS}, LDC\footnote{\url{https://www.ldc.upenn.edu}} and OpenSubtitle~\cite{opensubtitles2018}. The sizes and average length of these corpora are summarized in \Cref{tab:datasets_full}.


Below we review these document-level parallel corpora in detail. 
\paragraph{LDC}
This corpus consists of formal articles from the news and law domains.
The articles ave syntactic structures such as conjoined phrases, which make machine translation challenging.
However, the news articles in this corpus are relatively outdated.

\paragraph{IWSLT}
This corpus contains the TED Talks that covers the variety of topics. However, it is quite small in scale, which makes training large transformer-based models impractical.

\paragraph{News Commentary} 
This corpus consists of political and economic commentary crawled from the web site Project Syndicate\footnote{\url{4https://wit3.fbk.eu}}.  However, the scale of this corpus is also quite small. Moreover, there are no parallel Chinese-to-English data available in this corpus.

\paragraph{Europarl}
The corpus is extracted from the proceedings of the European Parliament. Only European language pairs are available in this corpus.

\paragraph{OpenSubtitle}
This corpus is a collection of translated movie
subtitles \cite{lison-tiedemann-2016-opensubtitles2016}. Besides being simple and short, the ``documents'' in this corpus are usually verbal and informal as well.

\paragraph{\BWB{}}
The \BWB{} corpus is the largest corpus in terms of size. 
Moreover, the sentences and documents in \BWB{} are substantially longer than previous corpora. It is also worth noting that \BWB{} differs from previous corpora in terms of genre -- in-depth human analysis shows that it is very challenging for current NMT systems due to its rich discourse phenomena.

\begin{table*}[htpb]
\begin{adjustbox}{width=0.99\textwidth}
{
\centering
\begin{tabular}{l|c|c|rrr|rrr}
\toprule[2pt]
\multirow{2}{*}{Corpus}  & \multirow{2}{*}{Genre}   & \multirow{2}{*}{Language} & \multicolumn{3}{c}{Size}                                                                         & \multicolumn{3}{c}{Averaged Length}                                                             \\
                         &                           &                           & \#word & \#sent & \#doc & \#word/sent & \#sent/doc & \#word/doc \\
\midrule[1pt]
	
\rowcolor{LightCyan}  \multirow{4}{*}{IWSLT~\cite{iwslt2020}}   & TED talk & ZH-EN                     & 4.2M                         & 0.2M                            & 2K                              & 19.5                           & 100                            & 2,100                         \\
                         &     TED talk        & FR-EN                     & 4.4M                         & 0.2M                            & 2K                              & 20.8                           & 100                            & 2,190                         \\
                         &    TED talk      & ES-EN                     & 4.2M                         & 0.2M                            & 2K                              & 19.9                           & 100                            & 2,080                         \\
                         &      TED talk     & DE-EN                     & 4.1M                         & 0.2M                            & 2K                              & 19.3                           & 100                            & 2,070                         \\
\midrule[1pt]
\multirow{2}{*}{NewsCom~\cite{tiedemann-2012-OPUS}} & News     & ES-EN                     & 6.4M                         & 0.2M                            & 5K                              & 30.7                           & 40                             & 1,288                         \\
                         &      News     & DE-EN                     & 6.4M                         & 0.2M                            & 5K                              & 33.1                           & 40                             & 1,288                         \\
\midrule[1pt]
Europarl~\cite{koehn-2005-europarl}                 & Parliament                & ET-EN                     & 7.3M                         & 0.2M                            & 15K                             & 35.1                           & 13                             & 485                           \\
\midrule[1pt]
\rowcolor{LightCyan}  LDC                     & News                      & ZH-EN                     & 81.8M                        & 2.8M                            & 61K                             & 23.7                           & 46                             & 1,340                         \\
\midrule[1pt]
\multirow{3}{*}{OpenSub~\cite{opensubtitles2018}} & Subtitle & FR-EN                     & 219.0M                         & 29.2M                           & 35K                             & 8.0                              & 834                            & 6,257                         \\
                         &   Subtitle     & EN-RU                     & 183.6M                       & 27.4M                           & 35K                             & 5.8                            & 783                            & 5,245                         \\
\rowcolor{LightCyan}             &      Subtitle     & ZH-EN                     & 16.9M                        & 2.2M                            & 3K                              & 5.6                            & 733                            & 5,647                         \\
\midrule[1.5pt]
\rowcolor{LightCyan}   \BWB{} (chapter)    &     Novel       & ZH-EN                     & \textbf{460.8M}                       & \textbf{9.6M}                            & \textbf{196K}                            & \textbf{48.1}                          & 49                             & 2,356                         \\
 \rowcolor{LightCyan} \BWB{} (book) &    Novel           & ZH-EN                     & \textbf{460.8M}                       & \textbf{9.6M}                            & 384                             & \textbf{48.1}                           & \textbf{25.0K}                          & \textbf{1.2M}          \\
\bottomrule[2pt]
\end{tabular}
}
\end{adjustbox}
    \caption{Statistics of various document-level parallel corpora. The parallel Chinese-English data is highlighted in \colorbox{LightCyan}{Cyan}.}
    \label{tab:datasets_full}
\end{table*}

\section{Case Study} \label{app:case_study}
We provide two example chapters in \BWB{} with coreference annotation in \Cref{fig:example_book1_0} and  \Cref{fig:example_book153_0}. 
We observe that the \BWB{} dataset poses challenges for NMT in the following ways.

\paragraph{Entity Consistency}
There are many named entities in the dataset that have a high repetition rate, such as fictional characters. 
Therefore, named entity consistency is a significant challenge in machine translation on this dataset.
For example, the translations of \orangeEnt{Weibo} and \redEnt{Qiao Lian} in \Cref{fig:example_book1_0} are not consistent in context.

\paragraph{Entity Recognition and Retrieval}
In addition to the fluency of entity translation, the adequacy of entity translation is another challenge in \BWB{}.
In the case of fictional characters with strange names, the NMT model may not correctly \emph{recognize} named entities, resulting in extremely poor translation quality, as in \Cref{fig:example_book153_0}. ``Ye Qing Luo'' could be literally translated as ``night'', ``clear'', ``fall''; however, it is actually a fictional characters.

Even though fictional characters are difficult to translate, they are relatively rare throughout the text, so it would be beneficial to abandon the assumption of inter-sentence independence in consideration of global contextual information.
One potential way to alleviate this problem is to equip NMT models with an entity recognition module.

\paragraph{Anaphoric Information Loss}
Chinese, being one of the pro-drop languages, omits many pronouns, while the English language does not, as shown in \Cref{tab:pronoun}. 
Translating from Chinese to English thus requires context to infer the correct English pronouns to compensate for the anaphoric information loss of sentence-level Chinese-to-English translations.

\paragraph{Morphological Information Loss}
Tense information is also frequently absent in Chinese and can only be inferred from context.
In general, this problem, which we refer to as \emph{morphological information loss}, is often encountered when translating from a morphologically poorer language to a morphologically richer one.
In the case of Chinese-to-English translation, tense information is often lost, while in other language pairs, such as English-to-French and English-to-German, gender information is often missed since as French and German are morphologically richer than English.

\paragraph{Coreference}
In addition, in \Cref{fig:example_book1_0}, we observe that the focus entity of the document is shifting throughout the text (\redEnt{Qiao Lian} $\xrightarrow{}$ \bleuEnt{Shen Liangchuan} $\xrightarrow{}$ \pinkEnt{Wang Wenhao} $\xrightarrow{}$ \bleuEnt{Shen Liangchuan} $\xrightarrow{}$ \cyanEnt{Song Cheng}), and this information is language-independent, i.e. consistent in source and target. This information could be used to improve the coherence of translation.

\begin{figure*}[t]
\begin{adjustbox}{width=\textwidth}
\centering \scriptsize
\begin{tabular}{p{5pt}p{150pt}p{150pt}p{150pt}}
\toprule[2pt]
& SOURCE & REFERENCE & MT \\
\midrule[1pt]
1) & \ZH{	\redEnt{乔恋}攥紧了拳头，垂下了头。	}&	\redEnt{Qiao Lian} clenched \redEnt{her} fists and lowered \redEnt{her} head.	&	\redEnt{Joe} clenched \redEnt{his} fist and bowed \redEnt{his} head.	\\
2) & \ZH{	其实\bleuEnt{他}说得对。	}&	Actually, \bleuEnt{he} \Verb{was} right.	&	In fact, \bleuEnt{he}\Verb{'s} right.	\\
3) & \ZH{自己就是一个蠢货，竟然会相信了网络上的爱情。	}&	\redEntOmit{She} \Verb{was} indeed an idiot, as only an idiot \Verb{would} believe that they could find true love online.	&	\redEntOmit{I} \Verb{am} a fool, even \Verb{will} believe the love on the Internet.	\\
4) & \ZH{	\redEnt{她}勾起了嘴唇，深呼吸一下，正打算将手机放下，微信上却被炸开了锅。	}&	\redEnt{She} curled \redEntOmit{her} lips and took a deep breath. Just when \redEntOmit{she} was about to put down \redEntOmit{her} cell phone, a barrage of posts bombarded \redEntOmit{her} WeChat account.	&	\redEnt{She} ticked \redEntOmit{her} lips, took a deep breath, and was about to put \redEntOmit{her} phone down, but weChat was blown open.	\\
5) & \ZH{	\redEnt{她}点进去，发现是\orangeEnt{凉粉群}，所有人都在@\redEnt{她}。	}&	\redEnt{She} logged into \redEntOmit{her} account and saw that a large number of fans in the \orangeEnt{Shen Liangchuan fan group} had tagged \redEnt{her}.	&	She nodded in and found it was a \orangeEnt{cold powder group}, and everyone was on \redEnt{her}.	\\
6) & \ZH{	【\redEnt{乔恋}：怎么了？	}&	[\redEnt{Qiao Lian}: What happened?]	&	\redEnt{Joe}: What's the matter?	\\
7) & \ZH{	【\grayEnt{川流不息}：\redEnt{乔恋}，快看\orangeEnt{微博}头条！ \orangeEnt{微博}头条？	}&	[\grayEnt{Chuan Forever}: \redEnt{Qiao Lian}, look at the headlines on \orangeEnt{Weibo}, quickly!]	&	\grayEnt{Chuan-flowing}: \redEnt{Joe love}, quickly look at the \orangeEnt{micro-blogging} headlines! \orangeEnt{Weibo} headlines?	\\
8) & \ZH{	\redEnt{她}微微一愣，拿起手机，登陆\orangeEnt{微博}，在看到头条的时候，整个人一下子愣住了！	}&	\redEnt{She} froze momentarily, then picked up \redEntOmit{her} cell phone and logged into \orangeEnt{Weibo}. When  \redEntOmit{she} saw the headlines,  \redEntOmit{her entire body} immediately froze over again!	&	\redEnt{She} took a slight look, picked up the phone, landed on the \orangeEnt{micro-blog}, when  \redEntOmit{she} saw the headlines,  \redEntOmit{the whole person} suddenly choked!	\\
9) & \ZH{	剧组发布会。 \bleuEnt{沈凉川}应邀出场，\grayEnt{导演}立马恭敬地迎接过来，客气的跟\bleuEnt{他}说这话，表达着\grayEnt{自己}对\bleuEnt{他}能够到来的谢意。	}&	\bleuEnt{Shen Liangchuan} arrived at the scene after accepting the invitation. \grayEnt{The director} immediately went to greet \bleuEnt{him} in a respectful manner, politely welcoming \bleuEnt{him} and expressing \grayEnt{his} gratitude for \bleuEnt{Shen Liangchuan}’s presence today.	&	The show's release. \bleuEnt{Shen Liangchuan} was invited to appear, \grayEnt{the director} immediately greeted \bleuEnt{him} with respect, politely said this to \bleuEnt{him}, expressed \grayEnt{his} gratitude for \bleuEnt{his} arrival.	\\
10) & \ZH{	对\bleuEnt{沈凉川}没有说话，看向不远处的\pinkEnt{王文豪}。	}&	\bleuEnt{Shen Liangchuan} did not speak. Instead \bleuEntOmit{he} looked at \pinkEnt{Wang Wenhao}, who was nearby.	&	\bleuEnt{Shen Liangchuan} did not speak, look not far from \pinkEnt{Wang Wenhao}.	\\
11) & \ZH{	\pinkEnt{王文豪}出事以后，所有的作品全部下架，而这一部剧还能播出，是因为\pinkEnt{王文豪}在里面友情饰演的男三号戏份很少，几乎可以忽略不计。	}&	After \pinkEnt{Wang Wenhao}’s scandal broke, every film \pinkEntOmit{he} starred in had been taken down. Only this show \Verb{could} still be broadcasted, as \pinkEnt{Wang Wenhao} \Verb{had} a supporting role in it and \Verb{was} practically unnoticeable.	&	After \pinkEnt{Wang Wenhao}'s accident, all the works were off the shelves, and this play \Verb{can} also be broadcast, because \pinkEnt{Wang Wenhao} in the friendship played by the male no. 3 play \Verb{is} very few, almost negligible.	\\
12) & \ZH{	剧组根本就没有邀请\pinkEnt{王文豪}，可\pinkEnt{他}却不知道从哪里拿到了邀请函，自己堂而皇之的进来了。 \pinkEnt{他}当然要进来了。	}&	In fact, the cast and crew hadn’t even invited \pinkEnt{Wang Wenhao}. However, \pinkEnt{he} had obtained a copy of the invitation letter somehow, and \Verb{strode} imposingly into the venue anyway.	&	The crew did not invite \pinkEnt{Wang Wenhao}, but \pinkEnt{he} did not know where to get the invitation, \pinkEnt{his} own entrance. Of course \pinkEnt{he}\Verb{'s} coming in.	\\
13) & \ZH{	这是\pinkEnt{他}最后的机会了。	}&	After all, this \Verb{was} \pinkEnt{his} final chance.	&	This \Verb{is} \pinkEnt{his} last chance.	\\
14) & \ZH{	丑闻闹出来，几乎所有的广告商和剧组都跟他毁约。	}&	After \pinkEntOmit{his} scandals broke, practically every advertiser and filming crew wanted to break their contracts with \pinkEnt{him}.	&	The scandal broke, and almost all advertisers and crews broke \pinkEntOmit{his} contract with \pinkEnt{him}.	\\
15) & \ZH{	\pinkEnt{他}现在宁可拍男三号，也不想就此沉寂。	}&	\pinkEnt{He} would rather take a supporting role than fade out into obscurity.	&	\pinkEnt{He} would rather shoot the men's number three now than be silent about it.	\\
16) & \ZH{	因为\pinkEnt{他}的事情，根本就压不下去。	}&	That was because the scandals surrounding \pinkEnt{him} \Verb{would} never disappear.	&	Because of \pinkEnt{his} affairs, there \Verb{is} no pressure.	\\
17) & \ZH{	所以\pinkEnt{王文豪}在发布会上，到处讨好别人。	}&	Thus, \pinkEnt{Wang Wenhao} went around trying to curry favor with everybody at this press conference.	&	So \pinkEnt{Wang Wenhao} tried to please others at the press conference.	\\
18) & \ZH{	\bleuEnt{沈凉川}穿着一身深灰色西装，面色清冷，手里端着一个高脚香槟杯，站在桌子旁边，整个人显得格外俊逸，却也格外的清冷，让周围的人都不敢上前搭讪。	}&	\bleuEnt{Shen Liangchuan} was wearing a dark grey suit and \bleuEntOmit{he} had a cold expression. \bleuEntOmit{He} was holding a champagne glass and was currently standing beside a table. \bleuEntOmit{He} looked exceptionally stylish, but also exceptionally icy. As a result, none of the people around \bleuEntOmit{him} dared to approach \bleuEntOmit{him}.	&	\bleuEnt{Shen 
River} was wearing a dark gray suit, \bleuEntOmit{his} face was cold, and \bleuEntOmit{he} was holding a tall champagne glass in \bleuEntOmit{his} hand. Standing beside the table, the whole person looked extraordinarily handsome, but also extraordinarily cold, so that people around \bleuEntOmit{him} did not dare to approach \bleuEntOmit{him}. 	\\
19) & \ZH{	可如果能注意到\bleuEnt{他}，就会发现\bleuEnt{他}的视线，却总是若有似无的飘到\pinkEnt{王文豪}身上。	}&	If anyone \Verb{had} paid attention to \bleuEnt{him}, they \Verb{would} have noticed that \bleuEnt{his} gaze \Verb{kept} drifting over to \pinkEnt{Wang Wenhao}.	&	But if you \Verb{can} notice \bleuEnt{him}, you \Verb{will} find \bleuEnt{his} sight, but always if there \Verb{is} nothing floating to \pinkEnt{Wang Wenhao} body.	\\
20) & \ZH{	\cyanEnt{宋城}站在\bleuEnt{他}的身边，察觉到这一点以后，就忍不住拽了拽\bleuEnt{他}的胳膊。	}&	\cyanEnt{Song Cheng} stood at \bleuEnt{his} side. After noticing \bleuEntOmit{his} behavior, \cyanEntOmit{he} \Verb{could} not help but pinch \bleuEnt{his} arm.	&	\cyanEnt{Songcheng} stood by his side, aware of this, \Verb{can} not help but pull \bleuEnt{his} arm.	\\
21) & \ZH{	\bleuEnt{沈凉川}淡淡回头，看向\cyanEnt{他}，目露询问。	}&	\bleuEnt{Shen Liangchuan} turned around and looked at \cyanEnt{him} casually, with a questioning face.	&	\bleuEnt{Shen Liangchuan} faint lying back, looked at \cyanEnt{him}, blind inquiry.	\\
22) & \ZH{	“\bleuEnt{沈哥}，您到底是要干什么啊？ 能不能告诉我，好让我有个心理准备。 您这样突然跑过来参加这么一个小剧组的发布会，又什么都不说就这么杵着，我心里瘆的慌。”	}&	“\bleuEnt{Brother Shen}, what are you planning to do? Can you tell me beforehand so that I can prepare myself mentally. You suddenly decide to come and attend such a small-scale press conference, yet you have been completely silent and are now just standing here and doing nothing? My heart is beating anxiously right now.”	&	\bleuEnt{Shen brother}, what the hell are you doing? Can you tell me so that I have a mental preparation. You suddenly ran over to attend the launch of such a small group, and said nothing so, I panicked. 	\\
23) & \ZH{	\bleuEnt{沈凉川}听到这话，抿了一口香槟，接着，将香槟杯放下。	}&	After \bleuEnt{Shen Liangchuan} heard him speak, he sipped a mouthful of champagne and put the glass down.	&	\bleuEnt{Shen} Said, took a sip of champagne, and then put the champagne glass down.	\\
24) & \ZH{	旋即，他迈开了修长的步伐。	}&	Then, \bleuEnt{he} walked away in long strides.	&	Immediately, \bleuEnt{he} took a slender step.	\\
25) & \ZH{	\cyanEnt{宋城}的心都提了起来，紧跟在\bleuEnt{他}身后。 \bleuEnt{沈凉川}一步一步往前，走到了前方。	}&	\cyanEnt{Song Cheng} was extremely nervous and followed \bleuEnt{him}. \bleuEnt{Shen} Liangchuan walked forward, one step at a time, until \bleuEntOmit{he} reached the front of the room.	&	\cyanEnt{Song Cheng}'s heart was raised and followed immediately behind \bleuEnt{him}. \bleuEnt{Shen Liangchuan} step by step forward, walked forward.	\\
26) & \ZH{	\pinkEnt{王文豪}正在跟别的三流小明星套近乎，那个人询问，“听说你打了一个狗仔？”	}&	\pinkEnt{Wang Wenhao} \Verb{was} currently ingratiating himself with a C-list celebrity. The celebrity asked, “Hey, I heard that you beat a paparazzi?”	&	\pinkEnt{Wang Wenhao} \Verb{is} with other third-rate star-studded sets, the man asked, "I heard you hit a paparazzi?" "	\\
27) & \ZH{	“对啊，现在的狗仔就是惹人厌恶，\pinkEnt{我}早就想动手教训他们了！ “\pinkEnt{你}这样，就不怕跟他们结仇啊？”	}&	“Yeah, the paparazzi nowadays are so disgusting. \pinkEnt{I} have wanted to teach them a lesson myself for some time now!” "Are not \pinkEnt{you} afraid of becoming an enemy of them?"	&	"Yeah, the paparazzi now are disgusting, \pinkEnt{I} have long wanted to teach them a lesson! "If \pinkEnt{you} are like this, \pinkEnt{you} are not afraid of enmity with them? "	\\
28) & \ZH{	''\pinkEnt{我}都这样了，\pinkEnt{我}怕什么？ 当初\bleuEnt{沈影帝}以正当防卫为借口，将一名狗仔打了，告到了法庭上去不也不了了之吗？``	}&	“\pinkEnt{I}’ve already done it, so what should \pinkEnt{I} be scared of? That time \bleuEnt{Best Actor Shen} beat up a reporter, \bleuEntOmit{he} claimed that it was in self-defence so that \bleuEntOmit{he} would have an excuse if he got sued, right? At that time, nobody said anything” 	&	"\pinkEnt{I}'m already like this, what am \pinkEnt{I} afraid of? Back then, \bleuEnt{Yingdi Shen} beat up a paparazzi under the pretext of self-defense, and sued \bleuEnt{him} in court, wouldn't it be over?	\\
29) & \ZH{	\pinkEnt{王文豪}说到这里，嘿嘿一笑。 还想说什么，忽然察觉到身后有人靠近。	}&	As \pinkEnt{Wang Wenhao} spoke, \pinkEntOmit{he} laughed heartily. Just as \pinkEntOmit{he} was about to continue speaking, \pinkEntOmit{he} suddenly felt a presence approaching \pinkEntOmit{him} from behind.	&	When \pinkEnt{Wang Wenhao} said this, \pinkEntOmit{he} smiled. \pinkEntOmit{He} was about to say something, but suddenly noticed someone approaching behind \pinkEntOmit{him}.	\\
30) & \ZH{	\pinkEnt{他}回头，就看到\bleuEntOmit{沈凉川}，眼瞳一缩，舔着脸笑，却见\bleuEntOmit{沈凉川}往前一步，一把揪住\pinkEnt{他}的衣领，接着一拳头对着\pinkEnt{他}的脸就砸过来！	}&	He turned around and saw \bleuEnt{Shen Liangchuan}. \bleuEntOmit{His} eyes narrowed and attempted to smile at \pinkEntOmit{him}. However, \bleuEnt{Shen Liangchuan} took a step forward, grabbed \pinkEnt{his} tie and threw a punch at \pinkEnt{his} face!	&	\pinkEnt{He} looked back, \pinkEntOmit{he} saw \bleuEnt{Shen Liangchuan}, eyes shrink, licking \bleuEntOmit{his} face and smiling, but saw Shen Liangchuan a step forward, a holding \pinkEnt{his} collar, and then a fist to \pinkEnt{his} face hit!	\\
\bottomrule[2pt]
\end{tabular}
\end{adjustbox}
\caption{An example chapter in \BWB. The same entities are marked with the same color. 
Pronoun omissions are marked with \Omit{}.
The mistranslated verbs are marked with \Verb{teal},
and the mistranslated named entities are marked with \orangeEnt{}.}
\label{fig:example_book1_0}
\vspace{-5pt}
\end{figure*}

\begin{figure*}[t]
\begin{adjustbox}{width=\textwidth}
\centering \scriptsize
\begin{tabular}{p{5pt}p{150pt}p{150pt}p{150pt}}
\toprule[2pt]
& SOURCE & REFERENCE & MT \\
\midrule[1pt]
1) & \ZH{	\redEnt{夜清落}浑身上下都传来剧烈的疼痛感，宛如千万把利刃，切割着\redEnt{她}的身体。	}&	\redEnt{Ye Qing Luo} suddenly felt an excruciating sharp pain tormenting \redEntOmit{her} entire body. It seemed as if a million sharp blades were slashing at \redEnt{her}.	&	\redEnt{The night fell} all over the body came a sharp pain, like a thousand sharp blades, cutting \redEntOmit{her} body.	\\
2) & \ZH{	尤其心脏那处，像是有着一团烈火，在体内燃烧，肆意的烧灼着\redEnt{她}的一切。	}&	\redEntOmit{Her} heart felt as if it was burning and that flame threatened to burn everything.	&	Especially the heart, like a fire, burning in the body, burning \redEnt{her} everything.	\\
3) & \ZH{	\redEnt{夜清落}想要伸手揉揉眉心，却发现自己全身虚软无力，连手指头都没法动弹。	}&	\redEnt{Ye Qing Luo} wanted to reach out but \redEntOmit{she} found that \redEntOmit{she} couldn’t move. \redEntOmit{She} felt so weak that \redEntOmit{she} could not even lift a finger.	&	\redEnt{Night clear} wants to reach out to rub the eyebrows, but found that \redEntOmit{their} whole body is weak, even fingers can not move.	\\
4) & \ZH{	耳边，隐隐约约传来模糊不清的声音。 “四少爷，这么做真的没事吗？” 	}&	In \redEntOmit{her} ears, there was a vague sound. “Fourth Young Master, are you sure it’s really alright to do this?”	&	in my ear, vaguely came a vague sound. "Four Masters, is it really okay to do this?"	\\
5) & \ZH{	“怕什么！”	}&	What are you so afraid of!	&	"Afraid of what! "	\\
6) & \ZH{	“太子殿下怎么可能看得上这个废物？ 要不是看她担着三泉宗少宗主的名头，太子殿下早就将她踹了。”	}&	How can His Royal Highness put such a good-for-nothing waste in his eyes? If not for her status as the Three Spring’s Lord’s daughter, do you think he would even bother with her?	&	"How could His Highness see this waste? Had it not been for her bearing the name of the Lord of Sanquan Zong, His Royal Highness would have taken her. "	\\
7) & \ZH{	“小废物可是夜四小姐亲自送来的，那碗药也是夜四小姐亲自喂的，太子殿下……说不定也是知情的。”	}&	She's a present personally sent to us by the Fourth Young Miss of the Ye family. That bowl of medicine was also personally fed to her by the Fourth Young Miss…so this matter…may also have been known by His Highness.	&	"Small waste is the night four Miss personally sent, that bowl of medicine is also the night four Miss personally fed, His Royal Highness ... ... Maybe it's also informed. "	\\
8) & \ZH{	妈哒！ 这是个什么鬼情况！	}&	Damn it, what kind of crappy drama is happening?	&	Damn! What a ghost situation!	\\
9) & \ZH{	\redEnt{夜清落}紧蹙着细眉，努力的睁开沉重的眼皮。	}&	\redEnt{Ye Qing Luo} scrunched \redEntOmit{her} brows together, mustering all her energy to lift her heavy eyelids.	&	\redEnt{The night fell} with a thin brow, and tried to open the heavy eyelids.	\\
10) & \ZH{	刚一睁开，就被极为耀眼的光芒，刺的她眼皮一痛。	}&	Just as \redEntOmit{she} opened them, \redEnt{her} eyes were stung by a bright light.	&	As soon as \redEntOmit{she} opened, she was the bright light, stabbing \redEnt{her} eyelids a pain.	\\
11) & \ZH{	一幕幕陌生的画面，宛如走马灯在脑海里不断的回旋。 	}&	Suddenly, \redEntOmit{her} mind reeled and it felt as if a there was an explosion in \redEntOmit{her} head. Fragments of unfamiliar pictures and scenes started to flood \redEntOmit{her} mind. 	&	A scene of strange scenes, like walking horse lights in the mind of the constant swing. 	\\
12) & \ZH{	斑斓画面一过，那些景象，像是强行插入的记忆，快速的在脑海里重叠，旋即渐渐归于平静。	}&	It continued to flash in \redEntOmit{her} mind non stop when \redEntOmit{she} suddenly realized that these fragments were forcing themselves into \redEntOmit{her} own memories as they melded and fused together. Soon, everything was calm.	&	The scene, those scenes, like forced insertion of memories, quickly overlapped in the mind, and gradually fell calm.	\\
13) & \ZH{	接收完这些记忆后，\redEnt{夜清落}再次睁开了眼睛。	}&	After \redEntOmit{she} received these memories, \redEnt{Ye Qing Luo} tried to pry open \redEntOmit{her} eyes again.	&	After receiving these memories, \redEnt{the night fell} and opened \redEntOmit{his} eyes again.	\\
14) & \ZH{	这一次，\redEnt{她}的眼睛适应了屋内的烛光摇动，灯火明耀。	}&	This time, \redEnt{her} eyes adapted quickly and focused on the candles.	&	This time, \redEnt{her} eyes adapted to the candlelight in the house, and the lights lit up.	\\
15) & \ZH{	奢华精致的房间，颇有古风意味，白色纱帐随风浮动。	}&	It was a luxurious room, exquisitely decorated in an ancient flavour.	&	Luxurious and sophisticated rooms, quite ancient, white yarn book with the wind floating.	\\
16) & \ZH{	四颗夜明珠伫立在房间四个角落，散发着莹莹的光芒。	}&	Four night pearls stood on each corner of the room, illuminating the room aglow.	&	Four night pearls stand in the four corners of the room, emitting a bright light.	\\
17) & \ZH{	最忒玛坑爹的是，\redEnt{她}现在四肢大敞的躺在一张圆木桌上，自己原本那具引以为傲的身材，变成了纤细娇弱的少女身体，只着了一件白色长袍。	}&	What seemed even more shocking was that \redEnt{she} was now lying atop a wooden table, \redEntOmit{her} voluptuous figure was now instead a slender and delicate girl's body, with only a white robe on.	&	The most beautiful father is that \redEnt{she} is now lying on a round wooden table with \redEntOmit{her} limbs, \redEntOmit{her} original proud figure, into a delicate girl's body, only a white robe.	\\
18) & \ZH{	靠！ \redEnt{她}这样躺在桌子上，是当做食物，等待着别人把\redEnt{她}吃掉吃掉吃掉咩？	}&	What the?! Lying on the table like this seemed to be as if \redEnt{she} was some kind of food and waiting for someone to eat \redEnt{her} up?	&	By! \redEnt{She} lay on the table like this, as food, waiting for others to eat \redEnt{her} and eat \redEnt{her}?	\\
19) & \ZH{	尼玛\redEnt{她}不就是从飞机上跳了下来，就把自己的灵魂转移到了这么一具小屁孩的身体里了？	}&	Damn it! \Verb{Didn’t} \redEnt{she} jump off the plane? How did her soul end up in this little girl’s body?	&	\Verb{Does} Neymar just jump off the plane and transfer her soul to such a little?	\\
20) & \ZH{	而且，刚穿越就这么惨，还能不能愉快的玩耍了？	}&	And since \redEntOmit{she}’s been given a new lease of life, \Verb{wasn’t} this start a bit too miserable?	&	And, just through such a miserable, \Verb{can} not be happy to play?	\\
21) & \ZH{	\redEnt{夜清落}想动，使不上劲。	}&	\redEnt{Ye Qing Luo} tried to move again, but \redEntOmit{she} could not muster any strength at all.	&	\redEnt{The night fell} and tried to move, so that \redEntOmit{it} could not be vigorous.	\\
22) & \ZH{	门外的那两人刚刚说过，\redEnt{她}被人下了药。	}&	As what the two men had said earlier, \redEnt{she} was drugged.	&	The two men outside the door had just said that \redEnt{she} had been drugged.	\\
23) & \ZH{	而且下药的那人，是这具身体的四姐，并且亲自把\redEnt{她}送到了这里，让人来玷污\redEnt{她}！	}&	And what’s more, the drug was personally administered by \redEntOmit{her} very own sister and even sent \redEnt{her} here as a present to these men to tarnish \redEnt{her}!	&	And the man who took the medicine, is the body of the four sisters, and personally sent \redEnt{her} here, let people to tarnish \redEnt{her}!	\\
24) & \ZH{	\redEnt{夜清落}快速从记忆中搜寻自己所需要的记忆。	}&	\redEnt{She} quickly searched through \redEntOmit{her} memories.	&	\redEnt{Night clearing} quickly searches for the memories \redEntOmit{you} need from memory.	\\
25) & \ZH{	门外那人，是玄者四大家之一\bleuEnt{尉迟}世家的四少爷尉迟涯，\bleuEnt{此人}风流成性，游手好闲，就是一个大写的纨绔少爷。	}&	The man who was outside, was from the \bleuEnt{Yuchi} clan, one of the four major family clans. \bleuEnt{He} was the Fourth Young Master of the \bleuEnt{Yuchi} family, a well known foppish playboy who spent, \bleuEntOmit{his} time idling about - \bleuEnt{Yuchi Ya}.	&	The person outside the door, is one of the four people of the \bleuEnt{Xuan}, the captain of the late family of the four young captain siaa, this person is a popular, idle, is a capital is a master.	\\
26) & \ZH{	把\redEnt{她}送到\bleuEnt{尉迟涯}面前，根本就是送羊入“狼”口！	}&	Sending her to \bleuEnt{Yuchi Ya} was simply putting a sheep in front of a wolf’s mouth!	&	Send \redEnt{her} to the captain in front of the \bleuEnt{late ya}, is simply to send sheep into the "wolf" mouth!	\\
27) & \ZH{	四姐？ 还有那个所谓的未婚夫？	}&	Fourth Sister? And who was her so called fiance?	&	Four sisters? And that so-called fiance?	\\
28) & \ZH{	呵！	}&	Ah!	&	Oh!	\\
29) & \ZH{	都给她等着！	}&	They better be good and wait for her to return this favour back many folds!	&	Just give her a wait!	\\
30) & \ZH{	\redEnt{夜清落}微眯起锋锐的眼睛，强压住身体传来的剧痛和麻木，努力的控制着四肢。	}&	\redEnt{Ye Qing Luo}'s gaze sharpened and she exerted a strong pressure, using all her effort to regain control of her limbs.	&	\redEnt{The night fell} slightly with sharp eyes, pressed the body from the sharp pain and numbness, and tried to control the limbs.	\\
31) & \ZH{	“吱呀”一声门响，\bleuEnt{尉迟涯}走了进来。	}&	[Squeak-] The door opened and \bleuEnt{Yuchi Ya} strode in.	&	"Squeaky" a door rang, \bleuEnt{the captain} came in late.	\\
32) & \ZH{	听脚步声，少说也有五人以上。	}&	From the sound of the footsteps, \redEntOmit{she} gathered that there were at least five or more people with \bleuEnt{him}.	&	Listen to the footsteps, less say there are more than five people.	\\
33) & \ZH{	“\redEnt{小废物}，\bleuEnt{哥哥}现在就来疼你！”	}&	\redEnt{Little Waste}, \bleuEnt{brother} is here to dote on you… 	&	"Little waste, \bleuEnt{brother} is here to hurt you now!" 	\\
34) & \ZH{	\bleuEnt{尉迟涯}走到桌子边，直接伸手扯\redEnt{她}身上的白袍。	}&	\bleuEnt{He} leered and slowly walked over to the table and immediately reached for \redEnt{her} white robe.	&	\bleuEnt{The captain} walked up to the table and reached directly for \redEnt{her} white robe.	\\
35) & \ZH{	\redEnt{夜清落}冰冷的眼神锐利，沙哑着嗓音，吐出一个字：“滚！” \bleuEnt{尉迟涯}听到\redEnt{她}的声音，笑得更是嚣张：“还没昏死过去？也好，也好！”	}&	When \bleuEnt{Yuchi Ya} heard \redEnt{her} voice, \bleuEntOmit{he} laughed even more lasciviously, with a hint of arrogance, "You’re awake? Very good, very good!"	&	\redEnt{The night clear} cold eyes sharp, hoarse voice, spit out a word: "Roll! When \bleuEntOmit{he} heard her voice, \bleuEntOmit{he} smiled more loudly: "Haven't passed out yet? Good, good!"	\\
\bottomrule[2pt]
\end{tabular}
\end{adjustbox}
\caption{Another example chapter in \BWB. This example is even more difficult for \MT{} since it fails to recognise the main character ``Ye Qing Luo'' as a named entity.}
\label{fig:example_book153_0}
\end{figure*}


\section{Experiment Setup} \label{app:model_parameters}
We adopt the parameters of Transformer Big \cite{transformer} for both \MTS\ and \MTD.
More precisely, the layers in the big encoders and decoders are $N=12$ , the number of heads per layer is $h = 16$, the dimensionality of input and output is $d_{model} = 1024$, and the inner-layer of a feed-forward networks has dimensionality $d_{ff} = 4096$. The dropout rate is fixed as 0.3. We adopt Adam optimizer with $\beta_1 = 0.9, \beta_2 = 0.98, \epsilon = 10^{-9}$, and set learning rate $0.1$ of the same learning rate schedule as Transformer. We set the batch size as 6,000 and the update frequency as 16 for updating parameters to imitate 128 GPUs on a machine with 8 V100 GPU. The datasets are encoded by BPE with 60K merge operations.

\section{The \BWB{} Corpus} \label{sec:datasetCreation}
In this section, we describe three stages of the dataset creation process: collecting bilingual parallel documents, quality control and dataset split.
\subsection{Bilingual Document Collection} \label{subsec:collection}
385 Chinese web novels across multiple genres were selected, including action, fantasy, romance, comedy, science fiction, martial arts, etc. The genre distribution is shown in \Cref{fig:genre_wordcloud}. 
We then scrape their corresponding English translations from the Internet.\footnote{\url{https://readnovelfull.com}}
The English versions are translated by professional translators who are native speakers of English, and then corrected and aligned by professional editors at the chapter level.
The text is converted to UTF-8 and certain data cleansing (e.g. deduplication) is performed in the process. 
Chapters that contain poetry or couplets in classical Chinese are excluded as they are difficult to translate directly into English.
Further, we exclude chapters with less than 5 sentences and chapters where the sequence ratio is greater than 3.0.
The titles of each chapter are also removed, since most of them are neither translated properly nor at the document level.
The sentence alignment is automatically performed by Bleualign\footnote{\url{https://github.com/rsennrich/Bleualign}}~\cite{bleualign}. 
The final corpus has 384 books with 9,581,816 sentence pairs (a total of 461.8 million words).\footnote{We will release a crawling and cleansing script pointing to a past web arxiv that will enable others to reproduce our dataset faithfully.}

\subsection{Quality Control} \label{subsec:quality}
We hired four bilingual graduate students to perform the quality control of the aforementioned process. %
These annotators were native Chinese speakers and proficient in English.
We randomly selected 163 chapters and asked the annotators to distinguish whether a document was well aligned at the sentence level by counting the number of misalignment. It is identified as a misalignment if, for example, line 39 in English corresponds to line 39 and line 40 in Chinese, but the tool made a mistake in combining the two sentences.
We observed an alignment accuracy rate of 93.1\%.

\subsection{Dataset Split} \label{subsec:split}
We construct the development set and the test set by randomly selecting 160 chapters from 6 novels, which contain 3,018 chapters in total.
To prevent any train-test leakage, these 6 novels are removed from the training set.
\Cref{tab:dataset_split} provides the detailed statistics of the \BWB{} dataset split.
In addition, we asked the same annotators who performed the quality control to manually correct misalignments in the development and test sets, and 7.3\% of the lines were corrected in total.

\section{Human Evaluation} \label{app:human_evaluation}
We conducted human evaluation on the \BWB{} test set following the protocol proposed by \citep{laubli-etal-2018-machine, laubli-etal-2020-set}. 
As stated in \Cref{sec:exp}, we evaluated two units of linguistic context (\Sentence{} and \Document{}) independently based on their respective \fluency{} and \adequacy{}.
We showed raters isolated sentences in random order in the \Sentence-level evaluation, whereas in the \Document-level evaluation, we presented entire documents and asked raters to evaluate a sequence of five sequential sentences at a time in order. 
The \adequacy{} evaluation was based solely on source texts, whereas neither source texts nor references were included in the \fluency{} evaluation.

The \adequacy{} evaluation was conducted by four professional Chinese to English translators, and the \fluency{} evaluation was conducted by four native English revisers.
The four translators were different from the professional translators who performed human translation. For human evaluation, we deliberately invited another group of specialists to avoid making judgments biased towards human translation. 

We adopted relative ranking because it has been shown to be more effective than direct assessment when conducted by experts rather than crowd workers~\cite{barrault-etal-2019-findings}.
In particular, raters were presented with the system outputs and were asked to evaluate the system outputs vis-à-vis one another, e.g. to decide whether system A was better than system B (with ties allowed). 

By randomizing the order of presentation of the system outputs, we were able to blind the origin of the output sentences and documents. While in the \Sentence-level evaluation, the system outputs were presented in different orders for each sentence, the \Document-level evaluation used the same ordering of systems within a document to help raters better assess global coherence.

Additionally, we used spam items for quality control.\cite{kittur2008crowdsourcing}. 
At the \Sentence-level, we make one of the five options nonsensical in a small fraction of items by randomly shuffling the order of the translated words, except for 10\% at the beginning and end. 
At the \Document-level, we randomly shuffle all translated sentences except the first and last sentence at the document level, rendering one of the five options nonsensical. 
If a rater marks a spam item as better than or equal to an actual translation, this is a strong indication that they did not read both options carefully.

Each raters evaluated 180 documents (including 18 spam items) and 180 sentences (including 18 spam items). The 180 sentences were randomly sampled from \testset{1} or \testset{2}.
We spited the test set into two non-overlapping subsets, referred to as \testset{1} and \testset{2}.
Note that \testset{1} and  \testset{2} were chosen from different books. 
Each rater evaluated both sentences and documents, but never the same text in both conditions so as to avoid repetition priming \cite{gonzalez2011cognitive}. 
Each document or sentence was therefore evaluated by two raters, as shown in \Cref{tab:human_evaluation_units}.

We report pairwise inter-rater agreement in \Cref{tab:kappa}. Cohen's kappa coefficients were used: 
\begin{equation}
    \kappa = \frac{P(A)-P(E)}{1-P(E)}
\end{equation}
where $P(A)$ is the proportion of times that two raters agree, and $P(E)$ is the likelihood of agreement by chance.

\begin{table}[]
\centering
\begin{tabular}{c|cc|cc}
\toprule[2pt]
 & \multicolumn{2}{c|}{\testset{1}} & \multicolumn{2}{c}{\testset{2}} \\
\adequacy  & \textsc{sent} & \textsc{doc}   & \textsc{sent} & \textsc{doc} \\ 
\midrule[1pt]
\rater{1} &             & \checkmark & \checkmark &             \\
\rater{2} &             & \checkmark & \checkmark &             \\
\rater{3} & \checkmark  &            &            & \checkmark  \\
\rater{4} & \checkmark  &            &            & \checkmark  \\ 
\bottomrule[2pt]
\end{tabular}
\begin{tabular}{c|cc|cc}
\toprule[2pt]
 & \multicolumn{2}{c|}{\testset{1}} & \multicolumn{2}{c}{\testset{2}} \\
\fluency  & \textsc{sent} & \textsc{doc}   & \textsc{sent} & \textsc{doc} \\ 
\midrule[1pt]
\rater{5} &             & \checkmark & \checkmark &             \\
\rater{6} &             & \checkmark & \checkmark &             \\
\rater{7} & \checkmark  &            &            & \checkmark  \\
\rater{8} & \checkmark  &            &            & \checkmark  \\ 
\bottomrule[2pt]
\end{tabular}
\caption{The evaluation units and corresponding raters. \rater{1-4} are professional Chinese to English translators and \rater{5-8} are native English revisers. }
\label{tab:human_evaluation_units}
\end{table} 
 
\begin{table}[]
\centering
\begin{tabular}{ccc}
\toprule[2pt]
              & \Sentence & \Document  \\ 
\midrule[1pt]
\rater{1}-\rater{2} & .171 & .169 \\
\rater{3}-\rater{4} & .294 & .346 \\
\rater{5}-\rater{6} & .323 & .402 \\
\rater{7}-\rater{8} & .378 & .342 \\ 
\bottomrule[2pt]
\end{tabular}
\caption{Inter-rater agreements measure by Cohen's $\kappa$.}
\label{tab:kappa}
\end{table}

\end{appendices}

\end{document}